\documentclass[times,twocolumn,final]{elsarticle}

\usepackage{medima}
\usepackage{framed,multirow}

\usepackage{latexsym}

\usepackage{amsmath,amssymb,amsfonts}
\usepackage{algorithmic}
\usepackage{graphicx}
\usepackage{textcomp}

\usepackage{microtype}
\usepackage[dvipsnames]{xcolor}
\usepackage{xspace}
\usepackage{hyperref}
\usepackage{cleveref}
\usepackage{wrapfig}
\usepackage{caption}

\newcommand{\algname}{ECAMP\xspace}

\makeatletter
\DeclareRobustCommand\onedot{\futurelet\@let@token\@onedot}
\def\@onedot{\ifx\@let@token.\else.\null\fi\xspace}
\def\eg{e.g\onedot}

\makeatother

\newcommand{\bheading}[1]{{\noindent{\textbf{#1}}}}

\newcommand{\br}{\mathbf{r}}

\makeatletter
\DeclareRobustCommand\onedot{\futurelet\@let@token\@onedot}
\def\@onedot{\ifx\@let@token.\else.\null\fi\xspace}
\def\eg{e.g\onedot}

\makeatother

\definecolor{darkgreen}{rgb}{0,0.7,0}
\definecolor{darkyellow}{rgb}{0.8,0.8,0}
\definecolor{bittersweet}{rgb}{1.0, 0.44, 0.37}
\definecolor{amber}{rgb}{1.0, 0.49, 0.0}
\definecolor{lgray}{rgb}{0.83,0.83,0.83}

\definecolor{color_unlabled}{rgb}{0.0,0.0,0.0}
\definecolor{color_vehicle}{rgb}{0.0,0.0,0.56}
\definecolor{color_road}{rgb}{0.5,0.25,0.5}
\definecolor{color_redlight}{rgb}{1.0,0.0,0.0}
\definecolor{color_person}{rgb}{0.859,0.078,0.234}
\definecolor{color_roadline}{rgb}{0.613,0.914,0.195}
\definecolor{color_sidewalk}{rgb}{0.953,0.137,0.906}

\definecolor{ellisred}{rgb}{0.87,0.44,0.38} 
\definecolor{ellisgreen}{rgb}{0.69,0.90,0.52} 
\definecolor{elliscyan}{rgb}{0.29,0.77,0.74} 
\definecolor{ellisorange}{rgb}{0.89,0.55,0.28} 
\definecolor{ellisblue}{rgb}{0.41,0.61,0.86} 

\newcommand{\cmtt}[1]{{\fontfamily{cmtt}\selectfont #1}}
\usepackage{multirow}
\usepackage{multicol}
\usepackage{colortbl}
\usepackage{tabularx}
\usepackage{tcolorbox}
\usepackage{etoolbox}
\usepackage{booktabs}
\usepackage{rotating}
\usepackage{soul}
\usepackage{cancel}

\usepackage{url}

\usepackage{hyperref}

\definecolor{color}{rgb}{.8,.349,.1}
\usepackage[dvipsnames]{xcolor}
\definecolor{skyblue}{RGB}{0, 100, 0}
\definecolor{gold}{rgb}{1, 0.843, 0}
\definecolor{coral}{rgb}{1.0, 0.498, 0.314}

\renewcommand{\arraystretch}{0.95}
\sethlcolor{yellow}

\journal{Medical Image Analysis}

\begin{document}

\verso{Wang, Yao, Jiang \textit{et~al.}}

\begin{frontmatter}

\title{ECAMP: Entity-centered Context-aware Medical Vision Language Pre-training}

\author[1,2,6]{Rongsheng Wang\fnref{fn1}}
\author[3]{Qingsong Yao\fnref{fn1}}
\author[1,2]{Zihang Jiang\corref{cor1}}
\author[1,2,6]{Haoran Lai}
\author[6]{Zhiyang He}
\author[6]{Xiaodong Tao}
\author[1,2,4,5]{S. Kevin Zhou\corref{cor1}}

\cortext[cor1]{Corresponding authors: skevinzhou@ustc.edu.cn (S Kevin Zhou) and  jzh0103@ustc.edu.cn (Z Jiang).}

\fntext[fn1]{Equal contribution.}
  
\address[1]{School of Biomedical Engineering, Division of Life Sciences and Medicine, University of Science and Technology of China (USTC), Hefei Anhui, 230026, China}
\address[2]{Center for Medical Imaging, Robotics, Analytic Computing \& Learning (MIRACLE), Suzhou Institute for Advance Research, USTC, Suzhou Jiangsu, 215123, China}
\address[3]{Stanford University, Palo Alto, California, 94025, United States}
\address[4]{Key Laboratory of Intelligent Information Processing of Chinese Academy of Sciences (CAS), Institute of Computing Technology, CAS}
\address[5]{Key Laboratory of Precision and Intelligent Chemistry, USTC, Hefei Anhui, 230026, China}
\address[6]{Anhui IFLYTEK CO., Ltd.}


\begin{abstract}
Despite significant advancements in medical vision-language pre-training, existing methods have largely overlooked the inherent linguistic complexity and imbalanced isssue within medical reports, as well as the complex cross-modality contextual relationships between texts and images. 
To close this gap, we propose a novel Entity-centered Context-aware Medical Vision-language Pre-training (\algname) framework, which establishes a more entity-centered, context-sensitive, and balanced understanding of medical reports to effectively pre-train the vision encoder. 
We first distill entity-centered context from medical reports utilizing large language models, enabling \algname to draw more precise supervision from the text modality. By further incorporating entity-aware re-balanced factor and descriptor masking strategies into masked languange modeling, 
\algname significantly enhances the knowledge of entities within the reports. 
A context-guided super-resolution task is proposed alongside a multi-scale context fusion design to improve the semantic integration of both coarse and fine-level image representations, which prompts better performance for multi-scale downstream applications. 
\algname integrates these innovations together, leading to significant performance leaps over current state-of-the-art methods and establish a new standard for cross-modality pre-training in medical imaging. The effectiveness of ECAMP is demonstrated by extensive experiments on various domains and organs, which achieves cutting-edge results on multiple tasks including classification, segmentation, and detection across 5 public chest X-ray and 4 fundoscopy datasets respectively.
\end{abstract}


\end{frontmatter}


\setlength{\skip\footins}{8 pt}


\section{Introduction}
\label{sec:intro}

By virtue of its exceptional accuracy and efficiency, deep learning (DL) is increasingly pivotal in medical image analysis~\cite{zhou2021review, pathak2022deep, esteva2017dermatologist, erickson2017machine, titano2018automated}. 
However, precision is contingent upon large-scale and high-quality annotations, which require an extensive number of experienced medical professionals, rendering it a costly endeavor.
 Fortunately, the diagnosis and experiential knowledge~\cite{pellegrini2023xplainer} of doctors are preserved in the form of paired reports and images, which provide valuable semantic context for training DL-based models.
More notably, these supervised signals are relatively accessible at little to no additional costs.

\begin{figure*}[t]
    \centering
    \includegraphics[width=\linewidth]{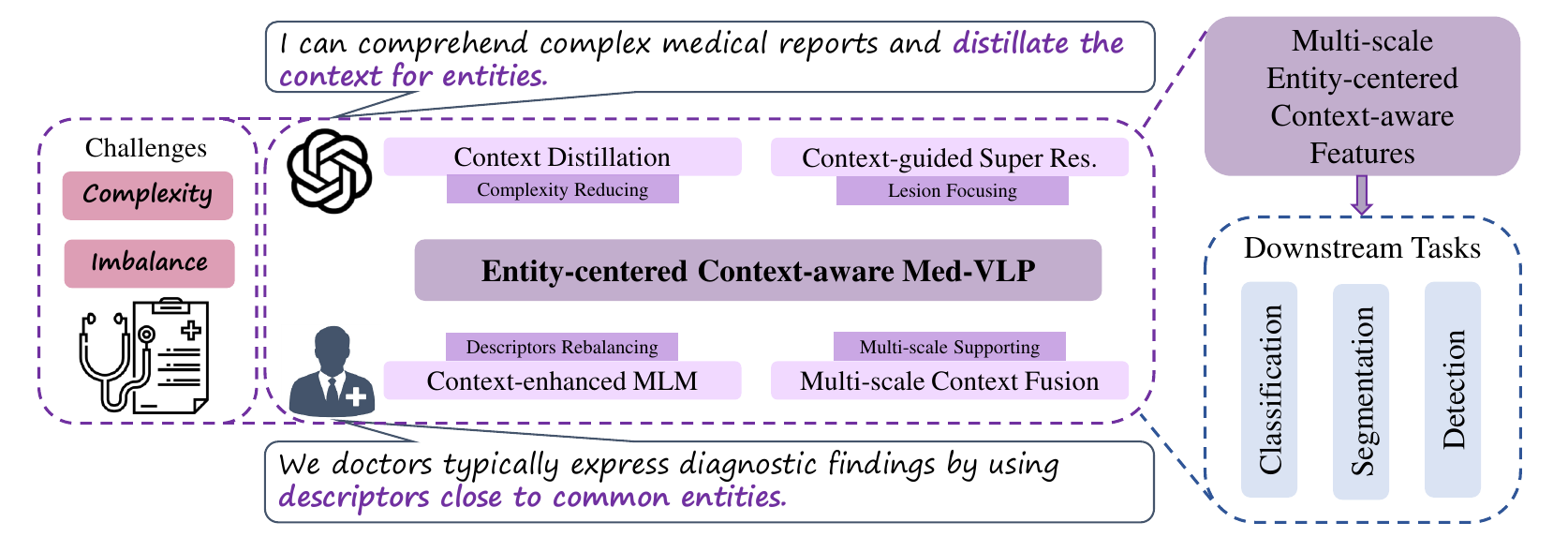}
    \vspace{-5mm}
    \caption{Our ECAMP is designed to learn representative and multi-scale features from complex and imbalanced medical reports, extracting medical knowledge from both ChatGPT and experts, with a particular focus on entities such as diseases, symptoms, and others. \textit{Context distillation} {prompts} ChatGPT to distillate precise content for entities, thereby reducing the complexity of reports, while \textit{Context-enhanced MLM} predicts the masked descriptors close to entities and addresses the imbalance difficulty caused by doctors' habits. \textit{Context-guided Super Resolution} focuses the model on anatomy area, concentrating the target pathology textures and locations. Further, ECAMP develops \textit{Multi-scale Context Fusion} to learn global and local projections simultaneously to support multi-granularity downstream tasks.}
    \label{fig:intro}
\end{figure*}

Cutting-edge algorithms~\cite{bannur2023learning, ye2024MedCoSS} in medical vision-language pre-training (Med-VLP) primarily leverage the information-rich medical report-image pairs to learn generic representations from two aspects: \textit{contrastive-based} methods align the text and image features at both global~\cite{convirt} and local~\cite{huang2021gloria,mgca,li2024mlip} levels; while \textit{reconstruction-based} methods generate the masked words~\cite{devlin2018bert} and image patches~\cite{He2021MAE} using cross-modality information~\cite{zhou2023mrm}. 

\begin{figure*}[t]
    \centering
    \includegraphics[width=1\linewidth]{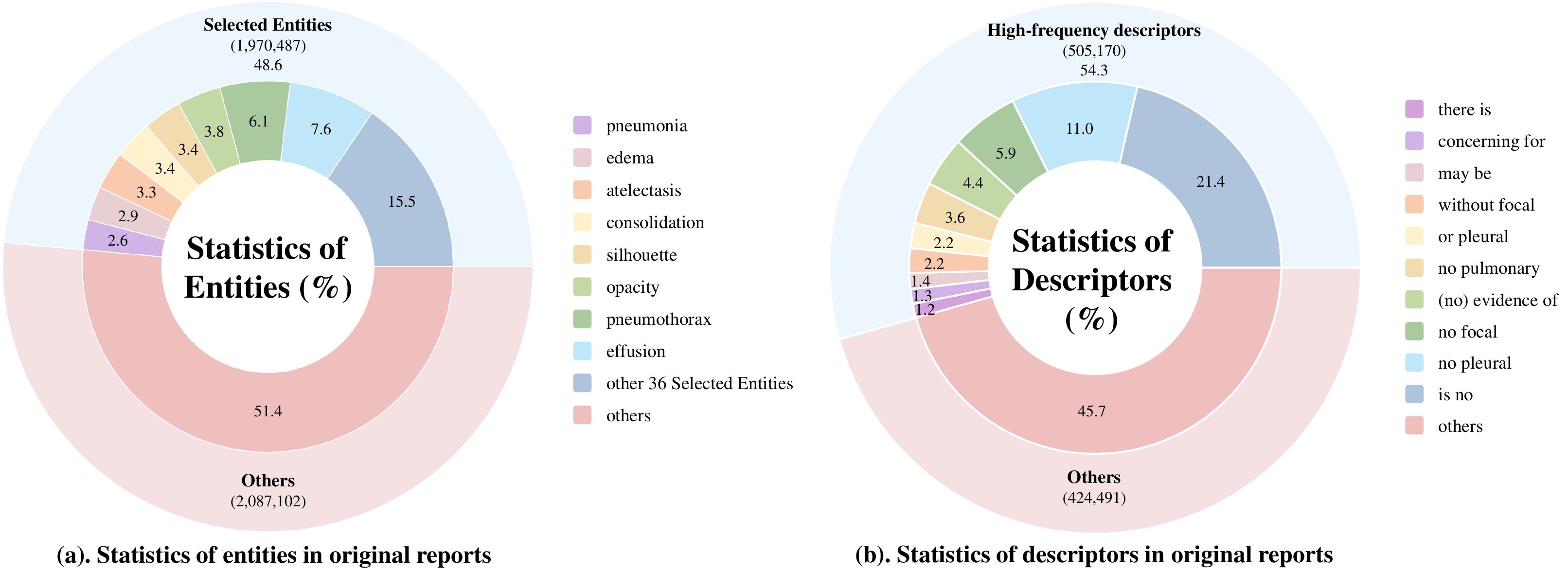}
    \vspace{-5mm}
    \caption{{The statistics of high-frequency entities and descriptors in the original reports of MIMIC-CXR}~\cite{johnson2019mimic}{. We merge other low-frequent words  as ``others''. Our selected entities account for approximately 48.6\% of the overall occurrences. The imbalance ratio between negative (e.g., ``is no'') and positive (e.g., ``there is'') descriptors reaches to 20:1.}}
    \label{fig:statistic}
\end{figure*}

Nevertheless, these methods still exhibit limitations in harnessing the potential of supervision in the context of the report. 
First, the linguistic challenge~\cite{li2019knowledge} posed by complex biomedical context~\cite{li2018hybrid} impedes the effectiveness of pre-training~\cite{Boecking2022BioVIL}.
Second, doctors often include diagnoses for common diseases (\eg, pneumothorax) in their reports, even when these diseases are negative~\cite{dai2021bdkg}. 
This practice leads to a significant context imbalance issue~\cite{Boecking2022BioVIL}: the majority of disease diagnoses are negative. We extract and count the two adjective words before disease entities (defined as ``descriptors'' ) and plot the statistics of these descriptors in Fig.~\ref{fig:statistic} (b)., which clearly indicate an imbalance: the  'is no' descriptor is dominant and the imbalance ratio between negative (e.g., ``is no'') and positive (e.g., ``there is'') descriptors reaches 20:1. This imbalance injects an adverse bias into the optimization of masked language modeling (MLM)~\cite{devlin2018bert} and complicates the model's ability to accurately learn the semantics of the positive diseases.

In this paper, we introduce \textbf{E}ntity-Centered \textbf{C}ontext-\textbf{A}ware \textbf{M}edical vision-language \textbf{P}re-training (\textbf{\algname}), a robust pre-training framework for 2D medical images.
As shown in Fig.~\ref{fig:intro}, the main contribution of ECAMP is pre-training representative and multi-scale features from complex and imbalanced medical reports through a cohesive integration of four synergistic modules. \algname consists of four simple-yet-effective components: 1) \textit{entity-aware context distillation}, 2) \textit{entity-centered context-enhanced MLM}, 3) \textit{context-guided super-resolution}, and 4) \textit{multi-scale context fusion}. We propose a cohesive integration of four modules, which contribute from different perspectives with close collaboration, emphasizing an entity-centered and context-aware design.

Unlike PRIOR~\cite{cheng2023prior} that substitutes complex biomedical sentences with prototypes, we propose \textit{entity-aware context distillation} to distill semantic context using Large Language Model, leveraging its strong power in linguistic comprehensive and summarization~\cite{gpt4}. 
Based on our consultations with medical professionals~\cite{dai2021bdkg}, doctors typically express diagnostic findings by using descriptors close to Medical Subject Headings (MeSH)~\cite{lipscomb2000MeSH} entities (\eg, pneumonia). Our key observations related to medical reports motivate the need for better distilling the semantic contexts related to clinically-relevant entities from reports.
Specifically, we use ChatGPT~\cite{brown2020gpt3} to summarize the complex reports into succinct expressions with entity diagnosis (like ``There may be pneumonia.''). We construct an entity list, including diseases and symptoms, and further generate more precise and simplified descriptions of disease existence and severity, which facilitate more effective supervision of the complex reports in pre-training. To the best of our knowledge, we are the first to leverage the power of large language models for medical vision-language pre-training, which attempts to explicitly address the challenge of information overload in clinically dense reports. The distilled context is not just a simplified report; it contains more precise descriptions of the existence and severity of diseases, enabling our model to learn more precise anatomical locations and disease-related features from the complex reports.

Despite that multi-modal MLM has been investigate in recent Med-VLP methods such as M3AE~\cite{chen2022m3ae} and MRM~\cite{zhou2023mrm}, they randomly mask portions of the report for construction, which overlook the importance and imbalance of the descriptive words beyond the entity. Accordingly, we propose \textit{entity-centered context-enhanced MLM} strategy. The \textit{descriptor masking} employs a fixed masking strategy to optimize the pre-trained model to predict these key descriptors for better modeling the semantics of entities and context.
To tackle the imbalance problem (as shown in Fig~\ref{fig:statistic} (b).), we further add an effective \textit{re-balancing factor} for less-appearing positive descriptors, which prevents the model from ignoring the critical positive diagnoses in reports.

Moreover, traditional MIM methods, such as MAE~\cite{He2021MAE}, treat each patch of the image equally during loss computation. However, lesions usually locate at small area of imaging. Super-resolution exhibits promise as a lesion focus and reinforcement of masked image modeling in self-supervised proxy task~\cite{zhou2023mrm}. Given that critical pathology usually occupies small patches, we hypothesize that learning to reconstruct these subtle cues help to better understand pathology and hence lead to better performance. 
Fortuitously, leveraging the concise distilled reports, we can retrain patch-level contrastive-based methods~\cite{huang2021gloria} to procure precise entity-specific attention maps. These maps facilitate our proposed \textit{context-guided super-resolution} to learn the high-resolution characteristics of the target pathology with broader contextual information, thereby enabling a more holistic and accurate interpretation of medical images based on the entities and context.
%

In addition, we find current cross-modal reconstruction-based methods typically employ either global average pooling (GAP)~\cite{zhou2023mrm} or cross-attention~\cite{chen2022m3ae} to integrate global and local vision representations to assist MLM.
In contrast, we introduce \textit{multi-scale context fusion} to aggregate global and local projection of pre-trained vision encoder simultaneously.
Multi-scale feature aggregation is definitely a trend for Med-VLP, but different methods has special designs which fit their frameworks. GLoRIA~\cite{huang2021gloria}, which uses contrastive learning to aggregate and align multi-scale features in two different branch. Instead, the only supervision which contains diagnosis information in our framework comes from masked report reconstruction, the gradient is propagated to local and global features via addition and duplication simultaneously, and train the multi-scale features.
Our motivation is to learn both global and local informative representations, which can be generalized to a wide range of multi-scale downstream tasks, including classification, segmentation, and detection.

\begin{figure*}[t]
    \centering
    \vspace{-3mm}
    \includegraphics[width=1\linewidth]{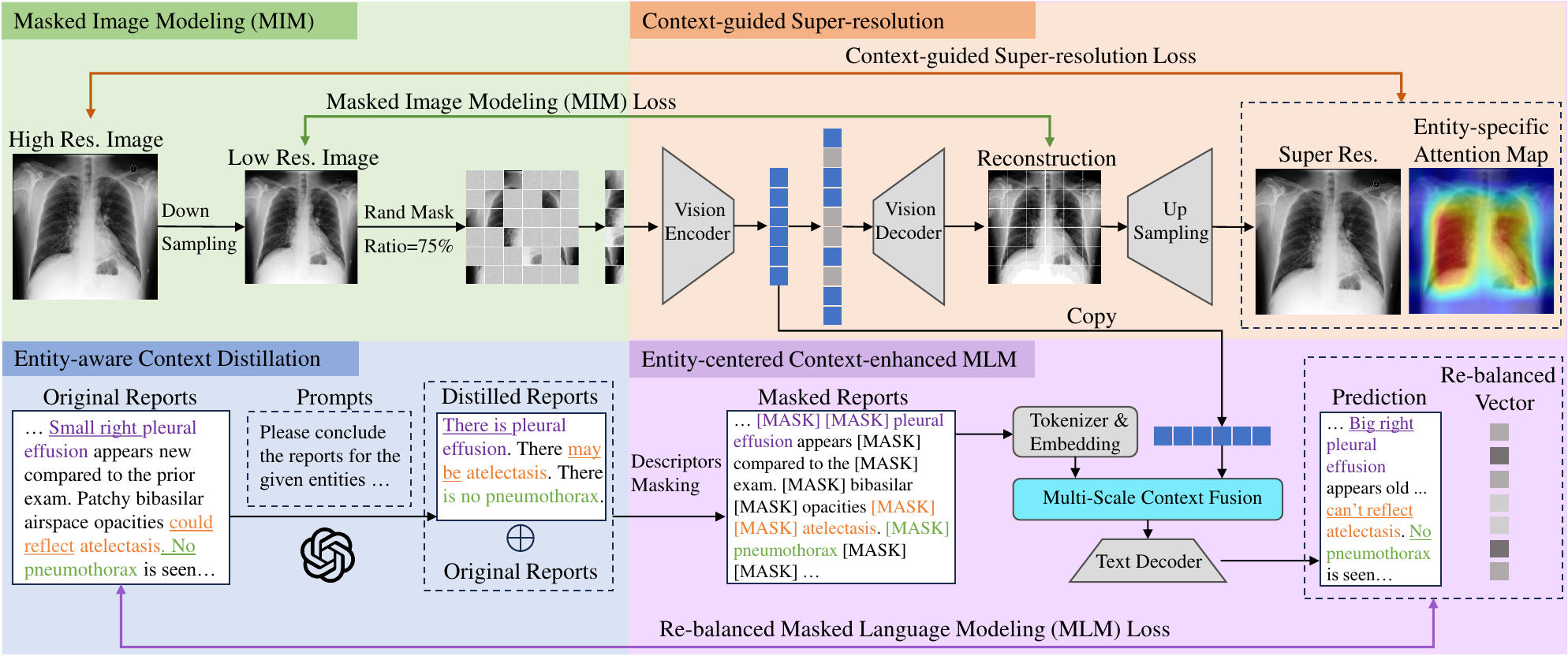}
    \vspace{-5mm}
    \caption{The overall framework of our \algname. Initially, \textit{entity-aware context distillation} succinctly encapsulates precise diagnosis for each entity. Then \textit{entity-centered context-enhanced MLM} masks the two descriptive words (underlined) preceding the entities, employing a re-balancing vector to direct the model's attention towards critical minority-positive expressions. To further enhance the learning of pathology vision contexts, \textit{context-guided super-resolution} utilizes an entity-specific attention map. Finally, we introduce a novel \textit{multi-scale context fusion} block (highlighted in \textcolor{cyan}{cyan}) to concurrently capture representative global and local features.}
    \label{fig:main}
    \vspace{-6mm}
\end{figure*}

To comprehensively and thoroughly validate the effectiveness, 
We pre-train our medical pre-training method \algname on MIMIC-CXR~\cite{johnson2019mimic} and FFA-IR~\cite{li2021ffa} dataset consisting of image-report pairs for both Chest X-rays and fundoscopies. Extensive experiments are conducted on multi-scale downstream tasks (including classification, segmentation, and detection) across multiple public datasets.
The experimental results  across various domains and organs demonstrate the superiority and generalization of \algname to outperform a plethora of competitive state-of-the-art (SOTA) Med-VLP methods with significant performance gaps. In sum, \algname with its readily accessible components prove valuable in fortifying the establishment of pre-training framework for medical foundation model.  Codes and models are available at \href{https://github.com/ToniChopp/ECAMP}{https://github.com/ToniChopp/ECAMP}.

%
%
%
%

\section{Related Work}

\subsection{General Vision Language Pre-training}

Vision-language pre-training (VLP) aims to improve the performance of multi-granularity downstream vision and language tasks by pre-training the model on large-scale image-text pairs.
Tremendous success has been achieved in recent literature, which can be summarized into two groups: encoder-based~\cite{li2021align,li2020unimo}, or encoder-decoder based~\cite{cho2021unifying, wangsimvlm} models. 
One of the most representative encoder-based methods is CLIP~\cite{clip}, which shows a great potential for learning mutual information between visual and linguistic data.
Recently, more works (e.g. GLIP~\cite{li2022glip}, SLIP~\cite{slip}, FLIP~\cite{flip}, OSCAR~\cite{li2020oscar}, and VinVL~\cite{zhang2021vinvl}) indicate that fine-grained context alignment facilitates the model to learn more representative representations.
For the encoder-decoder methods, BLIP~\cite{li2022blip} leverages a decoder to reconstruct text by leveraging vision semantic context.
In this paper, we focus on the challenging medical domain, where medical report is harder to comprehend and accuracy is more of paramount importance.

\subsection{Medical Vision Language Pre-training}

Recent Med-VLP methods can be categorized into two distinct types: report-supervised cross-modal alignment pre-training and reconstruction-based self-supervised pre-training. 
The groundbreaking works of the former type such as ConVIRT~\cite{convirt}, REFERS~\cite{zhou2022referes}, and CheXZero~\cite{tiu2022chexzero} pre-train the model by directly maximizing mutual information between the global representations.
GLoRIA~\cite{huang2021gloria}, SAT~\cite{liu2023sat}, and MGCA~\cite{mgca}  propose to align fine-grained features of paired image patches and words.
%
%
%
%
PRIOR~\cite{cheng2023prior} and BioVIL~\cite{Boecking2022BioVIL} attempt to comprehend intricate medical reports.
%
%
MedKLIP~\cite{wu2023medklip} and KAD~\cite{zhang2023kad} extract the medical-related information using a triplet extraction module as additional supervised signals.
%
FLAIR~\cite{silva2023flair} integrates the expert’s domain knowledge in the form of descriptive textual prompts.
RETFound~\cite{zhou2023RETFound} provides a basis for better label-effcient model adaptation.
Inspired by BERT~\cite{devlin2018bert} and MAE~\cite{He2021MAE}, the latter type learns representation by cross-modal context reconstruction task.
M3AE~\cite{chen2022m3ae} utilizes cross-attention for integrating multi-modal information to reconstruct masked tokens.
While MRM~\cite{zhou2023mrm} directly applies global average pooling for fusing visual features for masked text modeling.
Further, MPMA~\cite{zhang2023mpma} introduces a memory-augmented cross-modal fusion module to fully integrate visual information to assist report reconstruction.

Advanced efforts try to combine contrastive-based and reconstruction-based pre-training.
CMITM~\cite{chen2023cmitm} contrastively aligns global report and image representations after pre-training with MRM~\cite{zhou2023mrm}, while MedIM~\cite{xie2023medim} boosts MGCA~\cite{mgca} with attention guided masked image modeling.
Med-Unic~\cite{wan2023medUniC} and Med-MLLM~\cite{liu2023medmllm} further improve the performance by leveraging more pre-training datasets. 
Pair-Aug~\cite{xie2024pairaug} proposes pairwise augmentation to improve performance without introducing extra datasets.
Notwithstanding the enhancements, the aforementioned approaches are constrained by the intricacies of comprehending biomedical reports and the imbalance in diagnoses, which has spurred our pursuit of developing \algname.

\subsection{Large Language Models (LLMs)}

Compared to BERT~\cite{devlin2018bert}, recent LLMs such as GPT-3~\cite{brown2020gpt3} have a much larger scale in terms of training data and model parameters. This scale affords them robust zero-shot generalization capabilities to comprehend previously unseen contexts. Further, instruction-tuning~\cite{ouyang2022instructgpt,sanh2022multitask} is proposed to effectively improve the performance of LLM in novel application scenarios, including news summarization~\cite{zhang2023benchmarking} and code summarization~\cite{su2023distilled}. Luckily, advanced LLMs such as ChatGPT~\cite{brown2020gpt3} also exhibit robust capabilities in complex biomedical contexts. To the best of our knowledge, we are among the first few works to distill medical knowledge from medical reports using LLMs.

\begin{table}[t]
\small
\centering
\setlength{\tabcolsep}{3.8pt}
\caption{Our selected entities. We select 44 entities based on their importance and frequency of appearance in radiology reports as comprehensively as possible.}
\vspace{-3mm}
\begin{tabular}{llll}
\toprule
\multicolumn{4}{c}{\textbf{Entities}}\\\toprule                  
abnormality &abscess  &aerate  &aorta  \\
atelectasis &bronchiectasis  &calcification  &cardiomediastinal  \\
cardiomegaly  &catheter  &chf  &collapse  \\
congestion  &consolidation  &contour  &COPD  \\
deformity  &dilation  &distention  &edema  \\
effusion  &embolism  &emphysema  &engorgement  \\
fibrosis  &fracture  &granuloma  &hernia  \\
hilar  &hyperinflate  &hemidiaphragm  &infiltrate  \\
mass  &nodule  &obscure  &opacity  \\
perihilar  &pneumonia  &pneumothorax  &sarcoidosis  \\
silhouette  &thickening  &tuberculosis  &vasculature   \\\bottomrule
\end{tabular}
\label{table:entities}
\vspace{-5mm}
\end{table}

\section{Method}

In this section, we take CXRs and corresponding reports as an example to illustrate our proposed method. We first introduce how \algname leverages \textit{context distillation} to distill distinct context from LLM in \cref{sec:cd}. Then, we illustrate the basic masked image modeling (MIM)~\cite{He2021MAE} and the proposed \textit{entity-centered context-enhanced MLM} in \cref{sec:MM}. Next, \cref{sec:CgSR} shows the process of \textit{context-guided super-resolution}. Finally, we clarify the detailed approaches of \textit{multi-scale context fusion} in \cref{sec:MsCF} to integrate both global and local vision features for MLM. 

\vspace{-1mm}
\subsection{Context Distillation}
\label{sec:cd}

Given an image $I^o\in \mathbb{R}^{H^o\times W^o}$ and its original paired report $T_o$, our target is to extract concise diagnosis for diseases from $T_o$. Accordingly, we select the set of entities $R = \{r^1,r^2,\ldots,r^k\}$  from MeSH~\cite{lipscomb2000MeSH}, representing the diseases of interest in chest radiography. For each original report $T_o$, we identify entities $\br_o$ mentioned in the report through string matching. 
We manually curate a list of 44 entities, as shown in Table~\ref{table:entities}, including diseases, symptoms and anatomical locations. These entities are sourced from MeSH~\cite{lipscomb2000MeSH} and selected based on their importance and the frequency of appearance in radiology reports. As illustrated in Fig.~\ref{fig:statistic} (a)., the 44 entities selected for this study account for approximately 48.6\% of the overall occurrences.
Next, prompt $p$ is designed to require LLM to comprehend the complex biomedical context and summarize the diagnosis for disease entities  $\br_o$. 
Furthermore, to make the summarization of LLM stable, we leverage in-context learning~\cite{min2022incontext} to construct an one-shot~\cite{fewshotinstruct} sample $p'$ in the final prompt
$p=P(T_o, \br_o, p')$. 
Table~\ref{box:gpt_prompt} displays the prompt utilized for report distillation. By inputting the radiology report as the content to ChatGPT~\cite{gpt4} with this specific prompt, we obtain a distilled report that includes more detailed semantics regarding the existence and severity of entities. Specifically, by using GPT3.5-turbo API~\cite{gpt4} and setting temperature with 0 to generate 10,000 distilled reports, to save API call, we train a Vicuna7B~\cite{vicuna2023} to further distill all the reports. Examples of these distilled reports are presented in Table~\ref{table:distill_example}. \footnote{A medical student is tasked with evaluating the accuracy of the context distillation. The average distillation accuracy for ChatGPT 3.5 turbo is 93.6\%, while the accuracy for our Vicuna 7B is 93.2\%.}
The distilled report $T_d$ is finally computed as follows:
\begin{align}
    T_d = \mathrm{LLM}(P(T_o, \br_o, p')).
\label{eq:llm}
\end{align}
\subsection{Masked Image and Language Modeling}
\label{sec:MM}
\bheading{Masked image modeling}
We first decompose the original image $I^o$ into high-resolution $I^h\in \mathbb{R}^{H^h\times W^h}$ and low-resolution $I^l\in \mathbb{R}^{H^l\times W^l}$ input. 
The low-resolution image $I^l$ is split into $N_p^l$ patches with size $P \times P$.
Following MAE~\cite{He2021MAE}, we mask a large ratio ($75\%$) of the $N_p^I$ low-resolution patches, resulting in the unmasked patches $I_u=\{{I}_u^1,I_u^2, \ldots ,I_u^{N_u^I}\}$ and masked patches $I_m=\{{I}_m^1,I_m^2, \ldots ,I_m^{N_m^I}\}$.
Then the patch embedding $E_p$ is obtained by linearly projecting the flattened unmasked patches $I_u$.
Vision transformer~\cite{vit} (ViT) is chosen as the vision encoder $E_I$, which projects the patch embedding $E_p$ along with its positional embedding $E_{pos}$ as the vision feature $f_v = E_I(I_u, E_{pos})$. 
In the image decoding procedure, $f_v$ is fed into an image decoder $D_I$ after recovering to the original $N_p$ size, resulting in reconstructed image $\hat{I^l}=D_I(E_I(I_u))$.
We choose mean square error (MSE) loss to optimize the reconstructed image:
\begin{equation}
    \mathcal{L}_{\mathrm{MIM}}(\hat{I^l},I_m)=\mathrm{MSE}(D_I(E_I(I_u)), I_m).
\label{eq:mim}
\end{equation}

{
\captionsetup{type=table}
\captionof{table}{Our prompt used for report distillation is designed to include both an indication and an example. Providing ChatGPT~\cite{gpt4} with an example of the task helps enhance its performance in generating distilled reports.}\label{box:gpt_prompt}
\scriptsize
\begin{tcolorbox}[colback=gray!10,
                  colframe=black,
                  width=\linewidth,
                  arc=1mm, auto outer arc,
                  boxrule=0.5pt,
                 ]

{\cmtt{\textcolor{blue}{\textbf{Messages}}
 = [ 

\smallskip
 \{
"role": "system",
"content": f"""} \\ You are a knowledgeable and veteran doctor. \\
\cmtt{""" \},}

\smallskip
\cmtt{\{
"role": "user",
"content": f"""} \\
Please help me analysis the medical reports and conclude them briefly. Now, I will give you a medical report, as well as the mentioned entities. Please write brief and clear conclusions according to the reports as the following format: 'There is [a] [b].' when you are sure whether the entity exists in the report, or 'There may [a] [b].', when you are not sure whether the entity exists in the report. [a] represents adjective words describing the severeness or existence of the entity [b]. Please generate conclusions for the entities mentioned above one by one, according to the format and do not generate other words. Please keep only one entity in a sentence, there is no need of using 'and' or 'or' to connect two or more words. \\
\cmtt{""" \},}

\smallskip
\cmtt{\{
"role": "user",
"content": f"""}\\ ---Example----\\
Report: As compared to \_, the lung volumes have slightly decreased. Signs of mild over inflation and moderate pleural effusion persist. Elongation of the descending aorta.\\
Entities:  aorta, inflation, effusion\\
Conclusion: \\
There is moderate pleural effusion.\\
There is mild over inflation.\\
There is descending aorta. \\
---Example END----\\
Given the report:
\{\cmtt{\textcolor{blue}{report}\}}\\
Entities: \{\cmtt{\textcolor{blue}{entity}\}}.\\
Conclusion: \\
\cmtt{""" \},}]}

\end{tcolorbox}
}

\begin{figure}[t]
    \centering
    \includegraphics[width=0.75\linewidth]{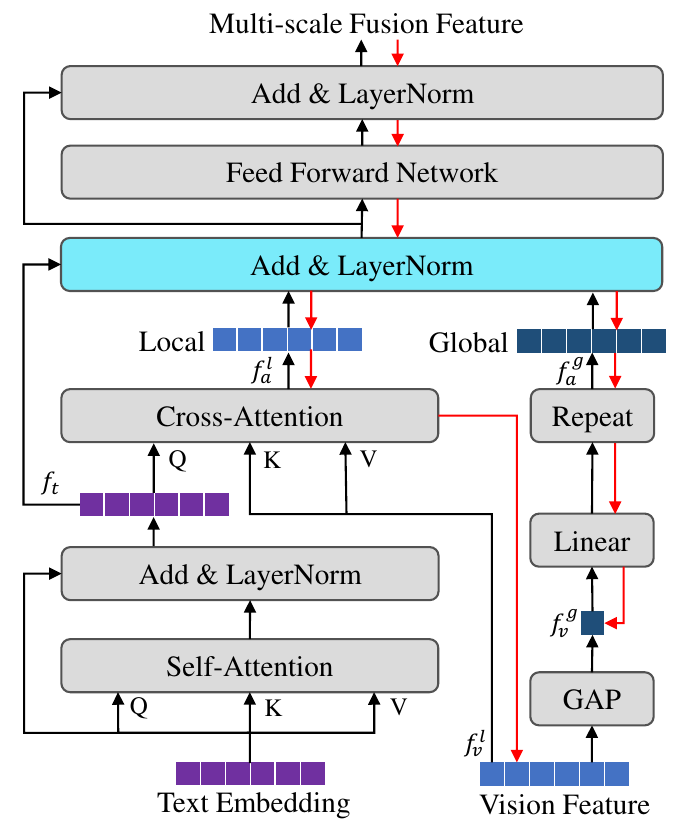}
    \vspace{-2mm}
    \caption{The framework of \textit{multi-scale context fusion}. The addition operation (highlighted in cyan) concurrently fuses the text feature, local aggregated vision feature, and global vision feature. The red arrows indicate the propagation of gradients, which simultaneously flow toward both global and local vision features.}
    \label{fig:gl-fusion}
\end{figure}

\bheading{Entity-centered context-enhanced MLM}
For the paired report $T_o$, we concatenate the distilled report $T_d$ as the text input $T = \left[T_o, T_d\right]$. 
As shown in Fig.~\ref{fig:statistic} (b)., we mark the first $\beta$ words preceding the entity $R$ as ``descriptive words'' $T_a=\{t_a^1,t_a^2,\ldots,t_a^{N^T_a}\}$  . Based on practical experience, the majority of descriptive words used by doctors are located within the first two words preceding the entity words, \eg, ``There \underline{is mild} pneumonia.'', we set $\beta$ to 2. 
Instead of using random masking~\cite{zhou2023mrm,chen2022m3ae,devlin2018bert}, \algname introduces a naive-yet-effective ``descriptors masking'' to mask the tokens of descriptive words, compelling the network to predict the diagnosis of the doctors for each entity. 
The remaining words are randomly masked with a ratio of $75\%$ to get masked words $T_m=\{t_m^1,t_m^2, \ldots ,t_m^{N_m^T}\}$ and unmasked words $T_u$, and the masked words are replaced with $\left[\text{MASK}\right]$ for training.
WordPiece~\cite{wordpiece} is chosen as tokenizer to convert $T$ into text tokens.%
Text embeddings $E_t$ are further computed by projecting the text tokens and adding the encoded position information.

\begin{table*}[h!]
\small
\centering
\setlength{\tabcolsep}{ 3pt}
\renewcommand{\arraystretch}{1.5}
\caption{{Examples of report distillation are presented. For each report, we showcase how the semantics of our selected entities are distilled by ChatGPT and our Vicuna-7B. The entities are highlighted in multiple colors for clarity. The distilled reports provide more specificity regarding the existence of entities.}}
\vspace{-3mm}
\begin{tabular}{l|p{16.5cm}}
\toprule
\multirow{3}{*}{\textbf{Original}}  &There is mild left base \textcolor{red}{atelectasis} seen on the frontal view without clear correlation on the lateral view. No definite focal \textcolor{purple}{consolidation} is seen. There is no pleural \textcolor{blue}{effusion} or \textcolor{cyan}{pneumothorax}. The \textcolor{green}{aorta} is slightly tortuous. The cardiac \textcolor{pink}{silhouette} is not enlarged. There is no overt pulmonary \textcolor{magenta}{edema}, mild left base \textcolor{red}{atelectasis}. otherwise, no acute cardio pulmonary process.  \\\hline
\multirow{2}{*}{\textbf{ChatGPT}}  &There is mild left base \textcolor{red}{atelectasis}. There is a slightly tortuous \textcolor{green}{aorta}. There is no pleural \textcolor{blue}{effusion}. There is no \textcolor{cyan}{pneumothorax}. There is no overt pulmonary \textcolor{magenta}{edema}. There is no acute cardio pulmonary process. \\\hline
\multirow{2}{*}{\textbf{Vicuna-7B}}  &There is mild left base \textcolor{red}{atelectasis}. There is no pleural \textcolor{blue}{effusion}. There is no \textcolor{cyan}{pneumothorax}. There is no definite focal \textcolor{purple}{consolidation}. There is no enlarged cardiac \textcolor{pink}{silhouette}. There is no overt pulmonary \textcolor{magenta}{edema}.  \\\hline
\hline
\multirow{3}{*}{\textbf{Original}}  &In comparison with the study of \_, the increased opacification at the right base has essentially cleared with better inspiration. Cardiac \textcolor{pink}{silhouette} remains at the upper limits of normal in size and there is again tortuosity of the \textcolor{green}{aorta} without vascular \textcolor{orange}{congestion} or pleural \textcolor{blue}{effusion}. Bi-apical changes especially on the right are stable.  \\\hline
\multirow{2}{*}{\textbf{ChatGPT}}  &There is tortuosity of the \textcolor{green}{aorta}. There is no pleural \textcolor{blue}{effusion}. There is no vascular \textcolor{orange}{congestion}. There is no change in the bi-apical changes. There is stable increased opacification at the right base. The cardiac \textcolor{pink}{silhouette} is at the upper limits of normal in size. 
 \\\hline
\textbf{Vicuna-7B}  &There is tortuosity of the \textcolor{green}{aorta}. There is no pleural \textcolor{blue}{effusion}. There is no vascular \textcolor{orange}{congestion}.\\\hline
\hline
\multirow{2}{*}{\textbf{Original}}  &Pa and lateral chest radiographs again demonstrate severe \textcolor{olive}{hyperinflation} and diffuse \textcolor{brown}{bronchiectasis}. There is no focal \textcolor{purple}{consolidation}, pleural \textcolor{blue}{effusion} or \textcolor{cyan}{pneumothorax}. The cardiomediastinal \textcolor{pink}{silhouette} is stable. No acute cardio pulmonary process. Findings were relayed to Dr \_.
  \\\hline
\multirow{2}{*}{\textbf{ChatGPT}}  &There is severe \textcolor{olive}{hyperinflation}. There is diffuse \textcolor{brown}{bronchiectasis}. There is no focal \textcolor{purple}{consolidation}. There is no pleural \textcolor{blue}{effusion}. There is no \textcolor{cyan}{pneumothorax}. The cardiomediastinal \textcolor{pink}{silhouette} is stable. There is no acute cardio pulmonary process. 
 \\\hline
\multirow{2}{*}{\textbf{Vicuna-7B}}  &There is severe \textcolor{olive}{hyperinflation}. There is diffuse \textcolor{brown}{bronchiectasis}. There is no focal \textcolor{purple}{consolidation}. There is no pleural \textcolor{blue}{effusion}. There is no \textcolor{cyan}{pneumothorax}.
  \\\hline
\hline
\multirow{3}{*}{\textbf{Original}}  &The lung volumes are normal. Mild \textcolor{skyblue}{cardiomegaly} which is stable. Normal \textcolor{gold}{hilar} and mediastinal structures. No \textcolor{coral}{pneumonia}. No pulmonary \textcolor{magenta}{edema} . No pleural \textcolor{blue}{effusion}. Status post cabg with aligned median stereotomy wires and normal location of surgical clips. Status post right lung surgery with surgical material seen. Mild \textcolor{skyblue}{cardiomegaly}. No evidence of \textcolor{coral}{pneumonia}.
  \\\hline
\multirow{2}{*}{\textbf{ChatGPT}}  &There is normal \textcolor{gold}{hilar}. There is mild \textcolor{skyblue}{cardiomegaly}. There is no \textcolor{coral}{pneumonia}. There is no pulmonary \textcolor{magenta}{edema}. There is no pleural \textcolor{blue}{effusion}. There is status post CABG with aligned median sternotomy wires. There is status post right lung surgery with surgical material. The lung volumes are normal.
 \\\hline
\multirow{2}{*}{\textbf{Vicuna-7B}}  &There is normal \textcolor{gold}{hilar}. There is mild \textcolor{skyblue}{cardiomegaly}. There is no \textcolor{coral}{pneumonia}. There is no pulmonary \textcolor{magenta}{edema}. There is no pleural \textcolor{blue}{effusion}.
  \\\hline\hline
\multirow{2}{*}{\textbf{Original}}  &3 radiographs are provided. On the film, 3 the tip of the dobbhoff \textcolor{violet}{catheter} projects over the proximal parts of the stomach. The course of the device is unremarkable. No complications. Unchanged appearance of the lung parenchyma the pleural and the heart.
  \\\hline
\multirow{2}{*}{\textbf{ChatGPT}}  &There is no complication with the Dobbhoff \textcolor{violet}{catheter}. There is unchanged appearance of the lung parenchyma. There is no pleural abnormality. There is no heart abnormality. There is no disease.
 \\\hline
\textbf{Vicuna-7B}  &There is no disease.
\\\bottomrule
\end{tabular}
\label{table:distill_example}
\end{table*}

A high text masking ratio challenges the model to reconstruct the reports by encoding the vision context thoroughly and accurately.
Accordingly, a novel \textit{multi-scale context fusion} module (MSCF, see \cref{sec:MsCF}) is proposed to integrate the vision feature and text embedding $f_f =  \mathrm{MSCF}(f_v,E_t)$.
The reconstructed text $\hat{T}$ is generated by decoding the fused feature $f_f$ using a BERT-based~\cite{devlin2018bert} text decoder $\hat{T} = D_t(f_f)$. The MLM loss for each masked word $t\in \left[T_m, T_a\right]$ is defined as $\mathcal{L}^t_{\mathrm{MLM}}(\hat{t}, t)=-\mathrm{log}(P(\hat{t}=t))$.

Moreover, we simply categorize $T_a$ into two groups: negative descriptions $T_{neg}$ and others $T_{oth}$, with $N^T_{neg}$ and $N^T_{oth}$ words, respectively.
Please note that since the expressions of positive descriptors are diverse, we divide them into others $T_{oth}$.
To better facilitate the pre-training model to learn precise disease semantics, we focus on modeling the minority-positive disease descriptions instead of majority-negative words, technically, we multiply the loss of $T_{neg}$ by a small number $\lambda_{neg}$, and the loss of $N^T_{oth}$ by $\lambda_{oth}$, where
\begin{align}
    \lambda_{neg}*N^T_{neg} + \lambda_{oth}*N^T_{oth}=N^T_a. 
\label{eq:neg}
\end{align}
%
According to the statistical results in Fig.~\ref{fig:statistic} (b)., the most negative descriptor ``there is no" occurs approximately $20\times$ more often than the most positive descriptor ``there may be.", so we set $\lambda_{neg}=0.05$.
Please note that $\lambda_{oth}$ is important to make sure that the model pays an enough attention to accurately reconstruct the critical ``descriptive words'' $T_a$. The final loss of entity-centered context-aware MLM is defined:
\begin{align}
    \mathcal{L}_{\mathrm{MLM}} = &\sum_j^{N^T_m}\mathcal{L}^t_{\mathrm{MLM}}(\hat{t}^j_m,t^j_m) +  \sum_j^{N^T_{neg}}\lambda_{neg}\mathcal{L}^t_{\mathrm{MLM}}(\hat{t}^j_{neg},t^j_{neg}) \nonumber\\ 
    &+ \sum_j^{N^T_{oth}}\lambda_{oth}\mathcal{L}^t_{\mathrm{MLM}}(\hat{t}^j_{oth},t^j_{oth}).
\label{eq:mlm}
\end{align}

\begin{table*}[t]
\small
\centering
\setlength{\tabcolsep}{ 7pt}
\caption{ Evaluating our method against other SOTA Med-VLP approaches on the fine-tuning classification task. $\dag$ denotes we reproduce results using their officially released models. Methods with \# leverage disease-level annotations. The best and second-best results are \textbf{bolded} and \underline{underlined}. }
\vspace{-3mm}
\begin{tabular}{l|ccc|ccc|ccc|ccc}
\toprule
\multirow{2}{*}{Method}        & \multicolumn{3}{c|}{ChestX-ray14 (AUC)}         & \multicolumn{3}{c|}{CheXpert (AUC)}  & \multicolumn{3}{c|}{RSNA (AUC)} & \multicolumn{3}{c}{COVIDx (ACC)}\\ \cline{2-13} 
  & \multicolumn{1}{c}{1\%} & \multicolumn{1}{c}{10\%} & \multicolumn{1}{c|}{100\%} & \multicolumn{1}{c}{1\%} & \multicolumn{1}{c}{10\%} & \multicolumn{1}{c|}{100\%} & \multicolumn{1}{c}{1\%} & \multicolumn{1}{c}{10\%} & \multicolumn{1}{c|}{100\%} & \multicolumn{1}{c}{1\%} & \multicolumn{1}{c}{10\%} & \multicolumn{1}{c}{100\%}\\ \toprule                  
Random init                      &60.0  &65.2  &72.8  &70.4  &81.1  &85.8  &71.9  &82.2  &88.5  &64.2  &75.4  &87.7\\
ImageNet init                    &69.8  &74.4  &80.0  &80.1  &84.8  &87.6  &83.1  &87.3  &90.8  &72.0  &84.4  &90.3\\\midrule
\multicolumn{1}{l}{\textit{CNN-based}} \\
GLoRIA~\cite{huang2021gloria}    &70.7  &78.6  &84.1  &86.6  &87.8  &88.1  &86.1  &88.0  &88.6  &73.3  &86.8  &90.5\\
$^{\dag}$SAT~\cite{liu2023sat}   &67.4  &79.3  &83.7  &86.3  &88.0  &88.2  &86.7  &87.5  &89.0  &75.0  &88.5  &94.3\\
PRIOR~\cite{cheng2023prior}      &75.7  &79.4  &84.3  &86.2  &88.3  &88.6  &85.7  &87.1  &89.2  &\underline{79.5}  &89.8  &94.8\\
MedKLIP\#~\cite{wu2023medklip}     &77.2  &78.9  &83.2  &87.1  &88.4  &88.6  &87.3  &88.0  &89.3  &78.3 &88.8 &95.3\\
KAD~\cite{zhang2023kad}          &78.7  &80.7  &82.5  &87.2  &88.2  &88.5  &89.8  &91.8  &92.5  &78.5 &90.5 &\underline{95.8}\\\midrule
\multicolumn{1}{l}{\textit{ViT-based}} & \\
MAE~\cite{He2021MAE}            &74.7  &81.3  &85.1  &80.7  &86.0  &86.7  &84.2  &89.6  &91.3  &69.8  &82.3  &90.8\\
GLoRIA~\cite{huang2021gloria}   &77.7  &82.8  &85.0  &86.5  &87.5  &87.8  &89.7  &91.2  &92.1  &76.8  &\underline{91.8}  &94.8\\
REFERS~\cite{zhou2022referes}    &76.7  &80.9  &84.7  &87.2  &88.1  &88.2  &89.4  &91.6  &92.7  &76.5  &90.8  &94.5\\
MGCA\#~\cite{mgca}                 &78.2  &82.7  &85.0  &87.0  &88.4  &88.5  &90.7  &92.6  &\underline{93.4}  &75.2  &91.5  &94.3\\
MRM~\cite{zhou2023mrm}           &\underline{79.4}  &\underline{84.0}  &\underline{85.9}  &\underline{88.5}  &\underline{88.5}  &\underline{88.7}  &\underline{91.3}  &\underline{92.7}  &93.3  &78.0  &90.3  &92.5\\
\rowcolor[gray]{0.9} \algname (Ours)      &\textbf{81.9}&\textbf{85.1}&\textbf{86.7}    &\textbf{88.8}&\textbf{89.5}&\textbf{89.7}    &\textbf{91.5}&\textbf{93.2}&\textbf{94.2}    &\textbf{83.0}&\textbf{95.5}&\textbf{97.5}\\\bottomrule
\end{tabular}
\label{Table:Finetune}
\end{table*}

\subsection{Context-guided Super-resolution}
\label{sec:CgSR}

Using the entity-centered distilled report $T_d$, we re-train the SOTA fine-grained representation contrastive learning method GloRIA~\cite{huang2021gloria} to generate entity-specific attention map $A$ for image $I^h$.
Then, an up-sampling module $D_{\mathrm{SR}}$ up-scales the reconstructed image $\hat{I^l}$ to the high-resolution results $\hat{I^h}$, which consists of a bi-linear interpolation and two convolution layers with a residual connection.
As the model should focus on generating disease-relevant visual textures, we use the attention map as the guidance for super-resolution (SR). The SR loss function is as follows:
\begin{equation}
    \mathcal{L}_{\mathrm{SR}}(\hat{I^h}, I^h)=\mathrm{MSE}(D_\mathrm{SR}(D_I(E_I(I_u))), I^h|A),
\label{eq:sr}
\end{equation}
The overall loss function of \algname is:
\begin{equation}
    \mathcal{L}=\mathcal{L}_\mathrm{MIM} + \mathcal{L}_\mathrm{MLM} + \mathcal{L}_\mathrm{SR}.
\label{eq:all}
\end{equation}

\subsection{Multi-scale Context Fusion}
\label{sec:MsCF}

Different from MRM~\cite{zhou2023mrm} which directly adds the vision feature $f_v$ to the text embedding $E_t$, we first apply a self-attention (SA) for an initial comprehension of text embedding $f_t = \mathrm{SA}(E_t)$.
In order to simultaneously attain representative global and local features for multi-scale downstream tasks, we leverage patch-wise vision features $f_v$ as local vision features $f_v^l$ and use global average pooling (GAP) to compute global vision feature $f_v^g = \mathrm{GAP}(f_v^l)$.
Next, we aggregate local vision context $f_v^l$ with text feature $f_t$ by cross-attention (CA) to generate local aggregated feature $f_a^l = \mathrm{CA}(f_v^l, f_t)$, and project the $f_v^g$ by a linear projector to obtain global feature $f_a^g$.
Finally, we integrate text feature $f_t$, local aggregated feature $f_a^l$, and duplicate global feature $f_a^g$ by addition. 
During the pre-training stage, the only supervision which contains diagnosis information in our framework comes from MLM loss (Eq.~\ref{eq:mlm}). So, as shown in Fig.~\ref{fig:gl-fusion},  the gradient is propagated to local vision feature $f_v^l$ and global vision feature $f_v^g$ via addition and duplication simultaneously, and train the multi-scale features.

\bheading{Fine-tuning stage:} \algname takes the full images as input without masking in this stage.
Our multi-scale context fusion design enables the model to learn compact global representation as well as informative fine-level local representation.
For the high-level downstream task like classification, we fine-tune the global vision feature $f_v^g$. 
For the low-level vision tasks including segmentation and detection, we fine-tune the local vision features $f_v^l$.
Thanks to the expressive representation, we are able to achieve impressive performance across different level vision tasks.

\vspace{-2mm}
\section{Experiment}

\begin{table*}[t]
\small
\centering
\setlength{\tabcolsep}{ 7pt}
\caption{ Comparison with other state-of-the-art methods on the linear probe classification task. $\dag$ means performances in ChestX-ray14~\cite{wang2017chestx} and methods like MRM~\cite{zhou2023mrm} are fine-tuned using their official released models. Methods with \# leverage disease-level annotations, while Med-UniC~\cite{wan2023medUniC} uses external additional pre-training data. The best and second-best results are \textbf{bolded} and \underline{underlined}.}
\vspace{-2mm}
\begin{tabular}{l|ccc|ccc|ccc|ccc}
\toprule
\multirow{2}{*}{Method}        & \multicolumn{3}{c|}{$\dag$ChestX-ray14 (AUC)}         & \multicolumn{3}{c|}{CheXpert (AUC)}  & \multicolumn{3}{c|}{RSNA (AUC)} & \multicolumn{3}{c}{COVIDx (ACC)}\\ \cline{2-13} 
  & \multicolumn{1}{c}{1\%} & \multicolumn{1}{c}{10\%} & \multicolumn{1}{c|}{100\%} & \multicolumn{1}{c}{1\%} & \multicolumn{1}{c}{10\%} & \multicolumn{1}{c|}{100\%} & \multicolumn{1}{c}{1\%} & \multicolumn{1}{c}{10\%} & \multicolumn{1}{c|}{100\%} & \multicolumn{1}{c}{1\%} & \multicolumn{1}{c}{10\%} & \multicolumn{1}{c}{100\%}\\  \toprule                  
Random init                      &52.1  &54.6  &55.3  &56.1  &62.6  &65.7  &58.9  &69.4  &74.1  &50.5  &60.3  &70.0\\
ImageNet init                    &67.0  &67.5 &71.6  &74.4  &79.7  &81.4  &74.9  &74.5  &76.3  &64.8  &78.8  &86.3\\\midrule
\multicolumn{1}{l}{\textit{CNN-based}} \\
SAT~\cite{liu2023sat}            &74.4  &79.2  &81.8  &86.9  &88.3  &88.6  &87.4  &89.2  &90.2  &74.5  &84.8  &89.3\\
$\dag$PRIOR~\cite{cheng2023prior}      &74.9  &79.5  &82.1  &86.3  &86.7  &87.3  &88.2  &90.4  &90.7  &74.8  &85.0  &89.8\\
MedKLIP\#~\cite{wu2023medklip}     &78.5  &80.9  &83.2  &86.2  &86.5  &87.7  &88.1  &90.8  &92.0  &74.5 &85.3 &90.3\\
$\dag$KAD~\cite{zhang2023kad}          &78.1  &79.8  &81.6  &86.4  &86.9  &87.8  &87.9  &90.0  &91.3  &75.0 &84.3 & 90.5\\
Med-UniC~\cite{wan2023medUniC}   &-  &-  &-  &88.2  &89.2  &89.5  &89.1  &90.4  &90.8  &76.5  &89.0  &92.8\\\midrule
\multicolumn{1}{l}{\textit{ViT-based}} & \\
MAE~\cite{He2021MAE}             &75.1  &79.7  &82.9  &82.4  &84.6  &85.2  &86.5  &89.7  &90.2  &79.0  &88.5  &92.5\\
GLoRIA~\cite{huang2021gloria}   &77.0  &81.9  &83.8  &84.6  &85.8  &86.2  &87.2  &88.1  &88.9  &73.3  &87.7  &92.1\\
MGCA\#~\cite{mgca}                 &78.7  &82.7  &84.1  &88.8  &89.1  &\underline{89.7}  &89.1  &89.9  &90.8  &74.8  &84.8  &92.3\\
$\dag$MRM~\cite{zhou2023mrm}           &\underline{78.8}  &\underline{82.7}  &\underline{84.6}  &88.1  &88.1  &88.3  &\underline{90.9}  &\underline{92.5}  &\underline{92.5}  &\underline{79.8}  &89.0  &92.8\\
MedIM-C~\cite{xie2023medim}        &-  &-  &-  &88.9  &89.3  &\underline{89.7}  &-  &-  &-  &77.2  &\underline{90.3}  &\underline{93.6}\\
MLIP\#~\cite{li2024mlip}  &-  &-  &-  &\underline{89.0}  &\underline{89.4}  &\textbf{90.0}  &89.3  &90.0  &90.8  &75.3  &86.3  &92.5\\
\rowcolor[gray]{0.9} \algname (Ours)      & \textbf{81.0}&\textbf{83.8}&\textbf{85.1}    &\textbf{89.3}&\textbf{89.5}&\underline{89.7}    &\textbf{91.4}&\textbf{92.9}&\textbf{93.5}  &\textbf{86.3}&\textbf{95.0}&\textbf{96.0}\\ \bottomrule
\end{tabular}.
\label{Table:linear}
\vspace{-3mm}
\end{table*}

\begin{table*}[t]
\small
\centering
\setlength{\tabcolsep}{12 pt}
\caption{Comparison with other state-of-the-art methods fine-tuned on semantic segmentation and object detection task. $\dag$ denotes performances reproduced using their officially released models. Methods with \# leverage disease-level annotations, while Med-UniC~\cite{wan2023medUniC} uses additional pre-training data. The best and second-best results are \textbf{bolded} and \underline{underlined}.}
\vspace{-2mm}
\begin{tabular}{l|ccc|ccc|ccc}
\toprule
\multirow{2}{*}{Method}  & \multicolumn{3}{c|}{SIIM (Dice)}  & \multicolumn{3}{c|}{RSNA (Dice)}  & \multicolumn{3}{c}{RSNA (mAP)}  \\
  & \multicolumn{1}{c}{1\%} & \multicolumn{1}{c}{10\%} & \multicolumn{1}{c|}{100\%} & \multicolumn{1}{c}{1\%} & \multicolumn{1}{c}{10\%} & \multicolumn{1}{c|}{100\%} & \multicolumn{1}{c}{1\%} & \multicolumn{1}{c}{10\%} & \multicolumn{1}{c}{100\%} \\  \toprule                  
Random init                      &9.0  &28.6  &54.3  &6.9  &10.6  &18.5  &1.0  &4.0  &8.9 \\
ImageNet init                    &10.2  &35.5  &63.5  &34.8  &39.9  &64.0  &3.6  &8.0  &15.7 \\\midrule
\multicolumn{1}{l}{\textit{CNN-based}} \\
MedKLIP\#~\cite{wu2023medklip}     &50.2  &60.8  &63.9  &66.2  &69.4  &71.9  &8.9  &16.3  &24.5 \\
KAD~\cite{zhang2023kad}          &35.8 &46.9 &63.4 &59.3 &67.5 &67.8 &9.8 &14.8 &18.8\\
SAT~\cite{liu2023sat}            &59.2  &68.2  &74.7  &67.8  &74.8  &75.3  &10.6  &19.1  &22.0 \\
PRIOR~\cite{cheng2023prior}      &54.6  &63.7  &76.8  &68.7  &71.9  &74.4  &13.8  &20.5  &25.6 \\
MGCA\#~\cite{mgca}                 &49.7 &59.3 &64.2  &63.0 &68.3 &69.8  &12.9 &16.8 &24.9 \\
Med-Unic~\cite{wan2023medUniC}   &56.7  &62.2  &64.4  &72.6  &74.4  &76.7  &16.6  &\underline{22.3}  &\underline{31.1} \\
MLIP\#~\cite{li2024mlip}           &51.6  &60.8  &68.1  &67.7  &68.8  &73.5  &\underline{17.2}  &19.1  &25.8 \\\midrule
\multicolumn{1}{l}{\textit{ViT-based}} & \\
$^{\dag}$MGCA\#~\cite{mgca}                 &62.7  &65.2  &71.4  &75.2  &\underline{77.8}  &\underline{78.3}  &8.9  &19.2  &26.3 \\
$^{\dag}$MRM~\cite{zhou2023mrm}           &\underline{63.1}  &\underline{68.3}  &78.4  &74.2  &77.6  &78.2  &11.5  &20.3  &27.1\\
Med-Unic~\cite{wan2023medUniC}            &62.1  &67.3  &71.5  &\underline{75.6} &76.6 &77.9 &-  &-  &- \\
MedIM-C~\cite{xie2023medim}        &-  &63.5  &81.3  &-  &-  &-  &-  &-  &- \\
MedIM~\cite{XIE2024MedIM}          &-  &64.2  &\underline{82.0}  &-  &-  &-  &-  &-  &- \\
\rowcolor[gray]{0.9} \algname (Ours)      & \textbf{67.3} &  \textbf{70.8}  & \textbf{84.5} & \textbf{77.1}   & \textbf{78.6}& \textbf{79.5} 
 & \textbf{19.8} &  \textbf{28.0}  &\textbf{31.6} \\ \bottomrule
\end{tabular}
\vspace{-2mm}
\label{table:seg-det}
\end{table*}

\subsection{Pre-training Dataset}
\bheading{MIMIC-CXR 2.0.0}~\footnote{https://physionet.org/content/mimic-cxr/2.0.0/}~\cite{johnson2019mimic,johnson2019mimicjpg} contains 377,110 chest radiographs and 227,835 corresponding reports. Each report may be associated with one or more images captured from frontal, back, and lateral views. 
We extract detailed descriptions of medical diseases by focusing on the impression and finding sections in the free-text reports.\\

\bheading{FFA-IR}~\footnote{https://physionet.org/content/ffa-ir-medical-report/1.0.0/}~\cite{li2021ffa} includes 10,790 reports along with 1,048,584 Fundus Fluorescein Angiography (FFA) images from clinical practice. It comprises explainable annotations, based on a schema of 46 categories of lesions; and it is bilingual, providing both English and Chinese reports for each case. We select English reports and paired FFA images as pre-training data. This setup is used to evaluate the performance of pre-training frameworks on fundoscopy images.

\subsection{Downstream Tasks Datasets}
\bheading{ChestX-ray14}~\footnote{https://nihcc.app.box.com/v/ChestXray-NIHCC}~\cite{wang2017chestx} comprises 112,120 frontal-view chest radiographs, and introduces a multi-label classification task involving \textit{14 common chest pathologies}. The dataset is officially split into training, validation, and testing subsets with a ratio of 70\%/10\%/20\%.

\bheading{CheXpert}~\footnote{https://stanfordmlgroup.github.io/competitions/chexpert/}~\cite{irvin2019chexpert} includes 223,648 chest radiographs from a frontal or lateral view, which categorizes each image into 5 diseases: atelectasis, cardiomegaly, consolidation, edema, and pleural effusion. Aligned with~\cite{convirt,huang2021gloria}, we designate the expert-labeled official validation set as the test data. We randomly choose 5,000 radiographs from the training data as the validation set. The split for training/validation/testing comprises 218,414/5,000/234 images.

\bheading{RSNA Pneumonia (RSNA)}~\footnote{https://www.rsna.org/rsnai/ai-image-challenge/rsna-pneumonia-detection-challenge-2018}~\cite{rsna} is composed of 29,700 frontal-view chest radiographs, each accompanied by pneumonia opacity bounding boxes if pneumonia is present in the image. Four tasks are conducted on this dataset: fine-tune classification, linear classification, semantic segmentation, and object detection. For fine-tune and linear classification tasks, we adhere to the official data partitioning, allocating 25,184, 1,500, and 3,000 images for the training, validation, and test sets, respectively. In the context of semantic segmentation and object detection, our approach aligns with the methodology proposed in~\cite{mgca}, wherein the original training set is randomly divided into subsets of 16,010/5,337/5,337 samples for training/validation/testing purposes, respectively.

\bheading{COVIDx CXR-3 (COVIDx)}~\footnote{https://www.kaggle.com/datasets/andyczhao/covidx-cxr2/versions/7}~\cite{wang2020covid} comprises 29,986 chest radiographs. This dataset presents a multi-class classification task, categorizing each image into \textit{COVID-19}, \textit{non-COVID pneumonia}, or \textit{normal}. In line with~\cite{mgca,wan2023medUniC,chen2023cmitm}, version 6 of this dataset is used. The official validation set is considered as our test data, and simultaneously, 10\% of the original training set is randomly selected as validation data. 

\bheading{SIIM-ACR Pneumothorax (SIIM)}~\footnote{https://www.kaggle.com/c/siim-acr-pneumothorax-segmentation}~\cite{siim} introduces the task of pneumothorax segmentation for evaluating the performance of segmentation models. This dataset includes 12,047 frontal-view chest radiographs with meticulously manually labeled pneumothorax masks. In accordance with~\cite{huang2021gloria}, we partition the dataset into training/validation/test sets with a ratio of 70\%/15\%/15\%.

\bheading{ODIR-5K}~\footnote{https://odir2019.grandchallenge.org/} consists of 5,000 patient records, each containing color fundus photographs of both the left and right eyes, along with corresponding diagnostic keywords provided by clinicians. These keywords are categorized into eight distinct classes.
We utilize the official test set as the test data. Additionally, 639 images are randomly selected as validation data, resulting in a data split of 70\% for training, 10\% for validation, and 20\% for testing.

\bheading{APTOS-2019}~\footnote{https://www.kaggle.com/datasets/mariaherrerot/aptos2019}~\cite{aptos2019-blindness-detection} presents a multi-class classification task for determining the severity level of diabetic retinopathy (DR) in fundus images. The dataset comprises 3,662 labeled images, with DR severity levels ranging from 0 to 4. The dataset is randomly partitioned into training, validation, and test sets with a split ratio of 70\%/10\%/20\%, respectively.

\bheading{MuReD}~\footnote{https://data.mendeley.com/datasets/pc4mb3h8hz/1}~\cite{MURED} consists of 2,451 multi-label retinal fundus images covering 20 categories. 
We randomly select 222 images from the official training set to serve as our validation data. The official test set is used for testing data. The data is split into 70\% for training, 10\% for validation, and 20\% for testing.

\bheading{RIGA}~\footnote{https://deepblue.lib.umich.edu/data/concern/data\_sets/3b591905z}~\cite{Ji_2021_MRNet} introduces the task of retinal cup and disc segmentation, containing a total of 750 color fundus images from three sources. We adopt the official test set, with the dataset split into 590 images for training, 95 for validation, and 95 for testing.

\subsection{Metrics}
\textbf{AUC and ACC} are used for multi-label and multi-class classification, respectively. Specifically, we calculate the average AUC for each class. 
\textbf{Dice} is commonly measured for segmentation tasks, we set the threshold to 0.5 as default for all experiments. 
\textbf{mAP} is adopted for object detection. Following ~\cite{mgca}, we set the IOU thresholds to 0.4, 0.45, 0.5, 0.55, 0.6, 0.65, 0.7, 0.75, and report the average precision.

\subsection{Implementation Details}
We utilize an 8-layer ViT-B/16~\cite{vit} as the backbone for the image encoder and a 4-layer ViT-B/16 for the image decoder. Additionally, we employ BERT~\cite{devlin2018bert} as the text encoder.
If not otherwise stated, we resize images to 224$\times$224 as input, except for super-resolution supervision we retain the 448$\times$448 resolution images. Texts are padded or truncated to a fixed length of 256 tokens. The latent dimension, denoted as D, is set to 768.
The implementation is based on PyTorch~\cite{paszke2019pytorch}.
For pre-training on MIMIC-CXR~\cite{johnson2019mimic,johnson2019mimicjpg}, our framework runs for 120 epochs on 4 A100 GPUs, taking approximately 20 hours. The training batch size is set to 256 for each GPU, and we utilize a gradient accumulation step of 8. Consequently, the effective batch size reaches 8,192. For pre-training on FFA-IR~\cite{li2021ffa}, our framework runs for 60 epochs on 4 A100 GPUs, taking approximately 15 hours. Other settings are the same with the pre-training stage on CXRs.
We opt for AdamW~\cite{loshchilov2017adamw} as our default optimizer, with a learning rate of 1.5e-4 and weight decay of 0.05. The values for $\beta_1$ and $\beta_2$ are set to 0.9 and 0.95. The mean squared error (MSE) loss is employed for masked image modeling and super-resolution, while cross-entropy loss is used for masked language modeling. 

For the LLM in Eq.~\ref{eq:llm}, we use GPT-3.5-turbo API~\cite{gpt4} with temperature setting to 0 to distill 10,000 reports and fine-tune a Vicuna model~\cite{vicuna2023} to distill the rest of the reports.

\begin{table*}[t]
\small
\centering
\setlength{\tabcolsep}{ 5.5pt}
\caption{ Ablation study of five simple-yet-effective components (\textit{context distillation (CD)}, \textit{descriptors masking (DM)}, \textit{re-balancing factor (RF)}, \textit{context-guided super-resolution (SR)}, and\textit{MSCF}) proposed in \algname on multi-scale downstream tasks including fine-tune classification (ChestX-ray14~\cite{wang2017chestx} ), linear classification (COVIDx~\cite{wang2020covid}), semantic segmentation (SIIM~\cite{siim}), and object detection (RSNA~\cite{rsna}). All of the downstream tasks are trained with \textbf{1\%}, \textbf{10\%} and \textbf{100\%} of training data.}
\vspace{-3mm}
\begin{tabular}{cccc|cc|ccc|ccc|ccc|ccc}
\toprule
\multirow{2}{*}{CD} & \multirow{2}{*}{DM}  &  \multirow{2}{*}{RF} & \multirow{2}{*}{SR} & \multicolumn{2}{c|}{MSCF}  & \multicolumn{3}{c|}{ChestX-ray14 (AUC) }   & \multicolumn{3}{c|}{COVIDx (ACC)}  & \multicolumn{3}{c|}{SIIM (Dice)}  & \multicolumn{3}{c}{RSNA (mAP)} \\ \cline{5-6}
&&& & Global & Local &1\% &10\% &{100\%} &1\% &10\% &{100\%} &1\% &10\% &{100\%} &1\% &10\% &{100\%} \\  \toprule
$\times$ & $\times$ & $\times$ & $\times$ & $\checkmark$ & $\checkmark$           &79.5  &83.9  &{86.0}  &78.0  &87.3  &{92.3}  &65.4  &68.6  &{80.7}  &4.1  &20.6  &{26.0}  \\
$\checkmark$ & $\times$ & $\times$ & $\times$ & $\checkmark$ & $\checkmark$       &80.2  &84.2  &{86.1}  &81.5  &89.0  &{89.5}  &65.9  &69.2  &{81.2}  &12.2  &22.4  &{26.6}  \\
$\checkmark$ & $\checkmark$ & $\times$ & $\times$ & $\checkmark$ & $\checkmark$   &80.8  &84.4  &{86.2}  &82.1  &90.3  &{92.8}  &66.0  &69.4  &{82.4}  &13.9  &23.3  &{27.3}  \\
$\checkmark$ & $\checkmark$ & $\checkmark$ & $\times$ & $\checkmark$ & $\checkmark$  &81.5  &84.7  &{86.5}  &83.0  &92.3  &{95.0}  &66.6  &69.9  &{83.6}  &16.1  &24.5  &{29.1}  \\
\hline
$\times$ & $\checkmark$ & $\checkmark$ & $\checkmark$ & $\checkmark$ & $\checkmark$  &81.3  &84.6  &{86.3}  &82.5  &92.5  &{95.0}  &66.9  &69.9  &{84.1}  &15.6  &25.2  &{29.7}  \\
$\checkmark$ & $\times$ & $\checkmark$ & $\checkmark$ & $\checkmark$ & $\checkmark$  &81.4  &84.5  &{86.2}  &83.0  &94.0  &{95.3}  &66.4  &70.0  &{83.4}  &16.8  &25.7  &{30.5}  \\
$\checkmark$ & $\checkmark$ & $\times$ & $\checkmark$ & $\checkmark$ & $\checkmark$  &80.8  &84.1  &{86.2}  &82.8  &91.8  &{92.5}  &66.6  &69.8  &{84.0}  &16.9  &26.5  &{30.3}  \\
\hline
$\checkmark$ & $\checkmark$ & $\checkmark$ & $\checkmark$ & $\checkmark$ & $\times$  &81.8  &85.0  &{86.4}  &85.0  &92.5  &{94.5}  &65.7  &69.5  &{83.6}  &12.6  &21.7  &{28.6}  \\
$\checkmark$ & $\checkmark$ & $\checkmark$ & $\checkmark$ & $\times$ & $\checkmark$  &81.4  &84.7  &{86.1}  &83.3  &88.0  &{90.8}  &\textbf{67.5}  &70.7  &{84.3}  &\textbf{20.4}  &\textbf{28.2}  &{\textbf{31.7}}  \\
\midrule
\rowcolor[gray]{0.9}  $\checkmark$ & $\checkmark$ & $\checkmark$ & $\checkmark$ & $\checkmark$ & $\checkmark$  &\textbf{81.9}  &\textbf{85.1}  &{\textbf{86.7}}  &\textbf{86.3}  &\textbf{95.0}  &{\textbf{96.0}}  &67.3  &\textbf{70.8}  &{\textbf{84.5}}  &19.8  &28.0 &{31.6}
\\ \bottomrule
\end{tabular}
\label{table:ab_component}
\end{table*}

\begin{table*}[t]
\small
\centering
\setlength{\tabcolsep}{ 8pt}
\caption{{Ablation study on the impact of two critical components: \textit{Entity Selection Strategy} and \textit{Text Input Composition}. We compare three entity selection strategies: (1) \textbf{Top-20} highest-frequency clinical entities, (2) \textbf{Top-10\%} frequency-percentile cutoff, and (3) our proposed method leveraging the \textbf{Top-44} most frequent entities. For text input composition, we assess four paradigms: (1) original radiology reports alone (\textbf{Origin-alone}), (2) ChatGPT-distilled reports alone (\textbf{Dis-alone}), (3) hybrid inputs combining original reports with named entity recognition (NER)-based distilled reports (\textbf{Hybrid-NER}), and (4) hybrid inputs combining original reports with ChatGPT-distilled reports (\textbf{Hybrid-GPT}). These configurations are evaluated on multi-scale downstream tasks including fine-tune classification (ChestX-ray14}~\cite{wang2017chestx}{), linear classification (COVIDx}~\cite{wang2020covid}{), semantic segmentation (SIIM}~\cite{siim}{), and object detection (RSNA}~\cite{rsna}{). All of the downstream tasks are trained with \textbf{1\%}, \textbf{10\%} and \textbf{100\%} of training data.}}
\vspace{-3mm}
\begin{tabular}{cc|ccc|ccc|ccc|ccc}
\toprule
{Entity}  &{Text Input}  & \multicolumn{3}{c|}{{ChestX-ray14 (AUC)}}   & \multicolumn{3}{c|}{{COVIDx (ACC)}}  & \multicolumn{3}{c|}{{SIIM (Dice)}}  & \multicolumn{3}{c}{{RSNA (mAP)}} \\
{Selection}  &{Composition}  &{1\%} &{10\%} &{100\%} &{1\%} &{10\%} &{100\%} &{1\%} &{10\%} &{100\%} &{1\%} &{10\%} &{100\%} \\  \toprule
{Top-20}  &{Hybrid-GPT}  &{81.2}  &{84.7}  &{86.4}  &{84.8}  &{92.8}  &{94.5}  &{66.7}  &{70.6}  &{82.5}  &{13.3}  &{25.3} &{26.9} \\
{Top-10\%}  &{Hybrid-GPT}  &{80.5}  &{84.0}  &{86.3}  &{83.5}  &{90.5}  &{93.3}  &{66.6}  &{70.6}  &{83.1}  &{17.1}  &{25.6}  &{28.0} \\
{Top-44}  &{Origin-alone}  &{81.3}  &{84.6}  &{86.3}  &{82.5}  &{92.5}  &{95.0}  &{66.9}  &{69.9}  &{84.1}  &{15.6}  &{25.2}  &{29.7} \\
{Top-44}  &{Dis-alone}   &{80.8}  &{84.2}  &{86.2}  &{82.0}  &{91.0}  &{94.5}  &{67.1}  &{69.6}  &{84.3}  &{15.3}  &{25.4} &{29.4} \\
{Top-44}  &{Hybrid-NER}   &{81.7}  &{84.9}  &{86.6}  &{84.0}  &{93.8}  &{95.3}  &{66.9}  &{70.1}  &{84.2}  &{16.2}  &{25.9} &{30.4} \\
\midrule
\rowcolor[gray]{0.9} {Top-44}  &{Hybrid-GPT}  &{\textbf{81.9}}  &{\textbf{85.1}}  &{\textbf{86.7}}  &{\textbf{86.3}}  &{\textbf{95.0}}  &{\textbf{96.0}}  &{\textbf{67.3}}  &{\textbf{70.8}}  &{\textbf{84.5}}  &{\textbf{19.8}}  &{\textbf{28.0}} &{\textbf{31.6}}
\\ \bottomrule
\end{tabular}
\vspace{-0.4cm}
\label{table:ab_text}
\end{table*}

\subsection{Baselines}

We conduct a comprehensive comparative analysis of \algname with numerous SOTA Med-VLP approaches \textit{pre-trained solely on MIMIC-CXR} using ResNet-50~\cite{resnet} (`CNN-Based') or ViT-B/16~\cite{vit} (`ViT-Based') as backbone, including MAE~\cite{He2021MAE} trained on ImageNet~\cite{krizhevsky2017imagenet} (denoted as `ImageNet init'), GLoRIA~\cite{huang2021gloria}, REFERS~\cite{zhou2022referes}, SAT~\cite{liu2023sat}, PRIOR~\cite{cheng2023prior}, MedKLIP~\cite{wu2023medklip}, MGCA~\cite{mgca}, KAD~\cite{zhang2023kad}, MedIM-C~\cite{xie2023medim}, MRM~\cite{zhou2023mrm}, MedIM~\cite{XIE2024MedIM} and MLIP~\cite{li2024mlip}. Notably, Med-UniC~\cite{wan2023medUniC} uses external multilingual datasets.
In addition, for fair comparison, we pre-train MAE~\cite{He2021MAE} (denoted as `MAE') and GLoRIA~\cite{huang2021gloria} with ViT-B/16 on MIMIC-CXR~\cite{johnson2019mimic} using official source code.
%
%
For MGCA, MRM, SAT, PRIOR, KAD, and MedKLIP whose official model are released, we fine-tune their models on the public downstream datasets in case there is no corresponding performance metrics in referenced papers.
For MedIM-C, Med-UniC, MedIM and MLIP, whose official models are absent, we directly copy the performances from their papers for comparison.
For extensive experiments on FFA images, we reproduce MAE~\cite{He2021MAE}, GLoRIA~\cite{huang2021gloria} and MRM~\cite{zhou2023mrm} using their official code pre-trained on FFA-IR~\cite{li2021ffa} dataset, which is under the same settings as our \algname.

\begin{figure}[h!]
    \centering
    \includegraphics[width=\linewidth]{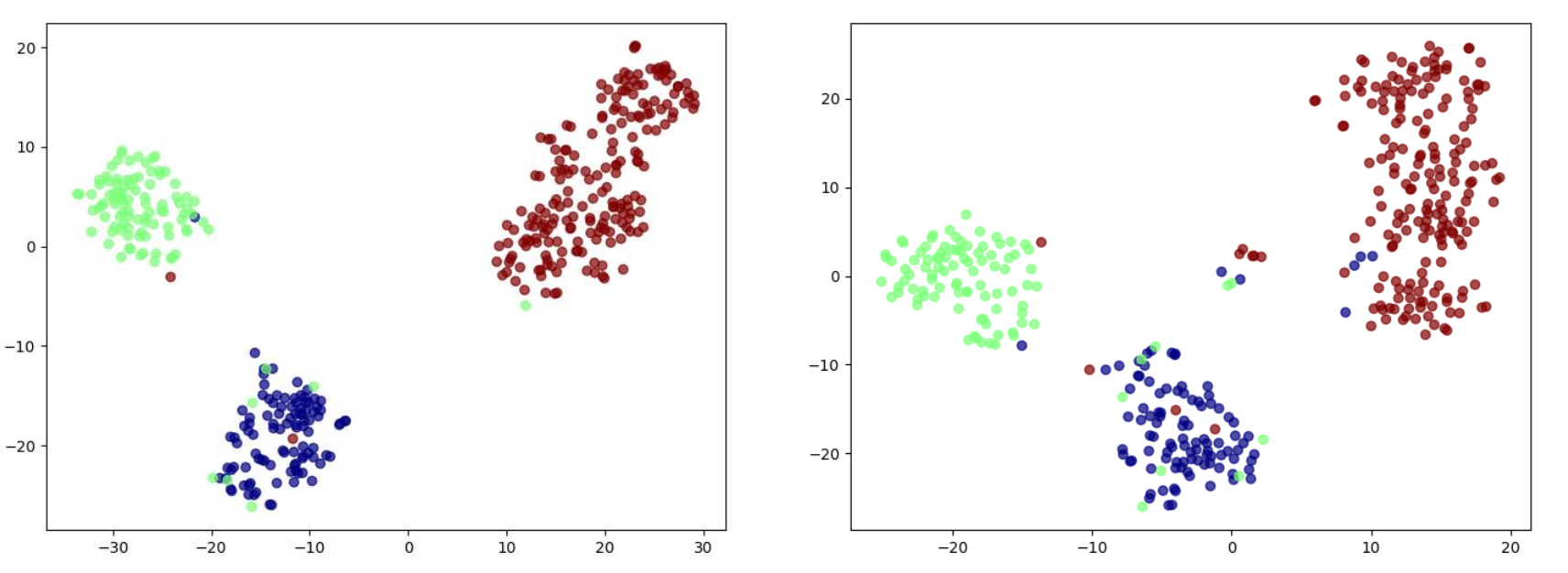}
    \caption{{T-SNE visualizations of ECAMP and MRM}~\cite{zhou2023mrm} {in the fine-tuning classification task on COVIDx}~\cite{wang2020covid} {dataset. The left panel represents ECAMP, while the right panel shows MRM. The {blue} points correspond to images labeled as “no pneumonia”, the {green} points represent image features related to “pneumonia”, and the {red} points indicate “COVID-19” features.}}
    \label{fig:tsne}
    \vspace{-5mm}
\end{figure}

\section{Results}
\subsection{Results on Classification}

We conduct both fine-tune and linear-probe with our \algname on four  
datasets, using 1\%, 10\%, and 100\%  of the training data, respectively.
%
As in Table~\ref{Table:Finetune}, our \algname significantly outperforms \textit{all} CNN and ViT-based competitive Med-VLP models across different ratios of training data on the four datasets. 

\begin{table}[t]
\small
\centering
\setlength{\tabcolsep}{ 8pt}
\caption{{Ablation study on parameter $\lambda_{neg}$ proposed in Eq.}~\ref{eq:neg} {on fine-tuning (ChestX-ray14}~\cite{wang2017chestx}{) and linear (COVIDx}~\cite{wang2020covid}{) classification with different ratios of training data (1\%, 10\%, and 100\%).}}
\vspace{-3mm}
\begin{tabular}{l|ccc|ccc}
\toprule
\multirow{2}{*}{$\lambda_{neg}$}  & \multicolumn{3}{c|}{ChestX-ray14 (AUC)}  & \multicolumn{3}{c}{COVIDx (ACC)}\\
  &\multicolumn{1}{c}{1\%}  &\multicolumn{1}{c}{10\%}  &\multicolumn{1}{c|}{{100\%}} &\multicolumn{1}{c}{1\%}  &\multicolumn{1}{c}{10\%}  &\multicolumn{1}{c}{{100\%}}\\ \toprule
0.025           &81.7  &84.5  &{86.5}  &85.0  &93.3  &{94.8}\\
\rowcolor[gray]{0.9}\textbf{0.05}   &\textbf{81.9}  &\textbf{85.1}  &{\textbf{86.7}}  &\textbf{86.3}  &\textbf{95.0}  &{\textbf{96.0}}\\
0.1             &81.8  &84.6  &{86.4}  &80.5  &94.8  &{95.3}\\
0.25            &81.5  &84.2  &{86.2}  &79.5  &94.0  &{94.5}\\
0.5             &81.4  &84.2  &{86.2}  &82.5  &91.5  &{91.8}\\\bottomrule
\end{tabular}
\vspace{-5mm}
\label{table:lambda}
\end{table}

It is worth noting that, on the ChestX-ray14~\cite{wang2017chestx}, \textit{which encompasses the most of testing data and disease labels, \algname surpasses the strongest counterpart by 2.5\% using only 1\% labeled data}.
%
Surprisingly, on the COVIDx with a novel disease ``COVID-19'', which is not present in MIMIC-CXR~\cite{johnson2019mimic}, a great performance gap can be found in Table~\ref{Table:Finetune} between \algname and SOTA methods. 
{COVIDx dataset includes the RSNA pneumonia dataset as well as a collection of chest X-rays featuring COVID-19 cases, which is similar to pneumonia. So, the pre-trained method has to learn different textures of pneumonia to further classify the COVID-19 of normal pneumonia. As shown in Fig.}~\ref{fig:tsne}{, we use t-SNE to visualize the features of our ECAMP and MRM. It is evident that our model demonstrates a better separation of features across “no pneumonia”; “pneumonia”, and “COVID-19”, which supports the superiority of pre-trained features of our ECAMP method and possibly enables better generalizability to novel class.}
This validates the superiority of the representation learned by our \algname.
Interestingly, we find the cross-modality reconstruction-based methods MRM~\cite{zhou2023mrm} and our \algname outperform the contrastive-based methods like MGCA~\cite{mgca} and GloRIA~\cite{huang2021gloria}.
%
Moreover, Table~\ref{Table:linear} presents the results of linear-probe classification. \algname exhibits consistently improved performance compared with SOTA methods on ChestX-ray14, RSNA and COVIDx datasets.
%
%
%
To evaluate the generalizability and adaptability of \algname, we extend the pre-training framework to Fundus Fluorescein Angiography (FFA) images and corresponding textual reports. Comprehensive experiments on fine-tuning classification is presented in Table~\ref{table:ffa-cls}. We perform fine-tuning for both multi-label and multi-class classification. As the table shows, our proposed \algname method substantially outperforms all state-of-the-art ViT-based medical vision-language pre-training approaches, even when applied beyond the CXR image domain. Notably, on the ODIR-5K dataset, which includes labels representing a broad spectrum of fundus diseases, \algname achieves a 5\% improvement over the strongest competing method, despite utilizing only 1\% of the labeled data.

\begin{table*}[t]
\small
\centering
\setlength{\tabcolsep}{14 pt}
\caption{ Comparison with other state-of-the-art medical vision language pre-training methods typically designed for CXRs pre-trained on FFA images and fine-tuned on disease classification tasks. $\dag$ denotes performances reproduced using their officially released codes pre-trained on FFA-IR~\cite{li2021ffa} dataset. The best and second-best results are \textbf{bolded} and \underline{underlined}.}
\vspace{-2mm}
\begin{tabular}{l|ccc|ccc|cc}
\toprule
\multirow{2}{*}{Method}  & \multicolumn{3}{c|}{ODIR-5K (AUC)}  & \multicolumn{3}{c|}{APTOS (ACC)}  & \multicolumn{2}{c}{MuRed (AUC)}  \\
  & \multicolumn{1}{c}{1\%} & \multicolumn{1}{c}{10\%} & \multicolumn{1}{c|}{100\%} & \multicolumn{1}{c}{1\%} & \multicolumn{1}{c}{10\%} & \multicolumn{1}{c|}{100\%} & \multicolumn{1}{c}{10\%} & \multicolumn{1}{c}{100\%} \\  \toprule                  
Random init                      &54.1  &57.5  &65.7  &59.7  &63.4  &68.5  &51.1  &66.5  \\
ImageNet init                    &62.6  &64.0  &77.4  &68.3  &71.7  &75.0  &52.9  &70.7 \\\midrule
\multicolumn{1}{l}{\textit{ViT-based}} & \\
$^{\dag}$MAE\cite{He2021MAE}                 &63.8  &72.9  &82.4  &71.8  &\underline{75.0}  &79.9  &68.8  &82.9  \\
$^{\dag}$GLoRIA\cite{huang2021gloria}              &\underline{67.4}  &74.1  &87.2  &72.3  &74.9 &\underline{81.0}  &81.0  &\underline{93.5}  \\
$^{\dag}$MRM\cite{zhou2023mrm}                 &66.8  &\underline{78.3}  &\underline{88.1}  &\underline{73.9}  &74.3  &80.9  &\underline{82.1}  &93.0  \\
\rowcolor[gray]{0.9} \algname (Ours)      & \textbf{71.8} &  \textbf{81.2}  & \textbf{90.0}  &\textbf{75.9}  &\textbf{76.9}  &\textbf{83.2}  &\textbf{85.6}  &\textbf{94.5} \\ \bottomrule
\end{tabular}
\vspace{-2mm}
\label{table:ffa-cls}
\end{table*}

\subsection{Results on Segmentation and Detection}

To assess the effectiveness of local representations learned by \algname, extensive experiments on fine-grained downstream tasks including segmentation and detection are conducted in Table~\ref{table:seg-det}.
%
Compared to strong SOTA approaches and boosted by MSCF, \algname makes remarkable improvements on all three datasets and ratios.
Even in scenarios where MGCA~\cite{mgca} and MedKLIP~\cite{wu2023medklip} leverage disease-level labels, and Med-UniC~\cite{wan2023medUniC} incorporates additional large-scale training data like PadChest~\cite{bustos2020padchest}, our \algname consistently outperforms all aforementioned methods. Distinguished Dice score leaps of 4.2\% and 1.5\% against SOTA methods are achieved on two segmentation datasets SIIM~\cite{siim} and RSNA~\cite{rsna} using 1\% training data, respectively. Notably, \algname reaches a Dice score of 67.3\%, which outperforms CNN-based methods fine-tuned using all training data on SIIM~\cite{siim}. Similarly, our method attains the best mAP using limited data on the object detection task.
Remarkably, our method exhibits a significant lead of 8.3\% compared to other SOTA open-source methods based on ViT.
These achievements substantiate the effectiveness of \algname to optimize informative local features for fine-grained downstream tasks. 
Table~\ref{table:ffa-seg} presents the results of our fine-grained downstream task for retinal cup and disc segmentation on RIGA~\cite{Ji_2021_MRNet} dataset when pre-trained on FFA images and paired reports. Here, the average Dice coefficient is recorded as the primary metric. Our proposed method, \algname, consistently outperforms all previously mentioned approaches on this task, demonstrating its superior performance.
Through extensive experiments conducted on FFA images, we have validated that \algname effectively leverages its strong capability to utilize contextual information and capture multi-granularity cross-modality relationships between text and images. This enables \algname to generalize well across different medical vision tasks, making it a robust and highly effective medical vision-language pre-training method.

\begin{table}[t]
\small
\centering
\setlength{\tabcolsep}{ 10pt}
\vspace{-2mm}
\caption{Comparison with other SOTA medical vision language pre-training methods pre-trained on FFA images and fine-tuned for retinal cup and disc segmentation task. The average Dice of retinal cup and disc is shown. $\dag$ denotes performances reproduced using their officially released codes pre-trained on FFA-IR~\cite{li2021ffa} dataset. The best and second-best results are \textbf{bolded} and \underline{underlined}. }
\vspace{-2mm}
\begin{tabular}{l|ccc}
\toprule
\multirow{2}{*}{Method}  & \multicolumn{3}{c}{RIGA (Dice)}  \\
  & \multicolumn{1}{c}{1\%} & \multicolumn{1}{c}{10\%} & \multicolumn{1}{c}{100\%} \\  \toprule                  
Random init                      &50.9&65.0&78.3  \\
ImageNet init                    &63.3&73.8&82.5 \\\midrule
\multicolumn{1}{l}{\textit{ViT-based}} & \\
$^{\dag}$MAE~\cite{He2021MAE}                &67.0&78.1&86.4  \\
$^{\dag}$GLoRIA~\cite{huang2021gloria}          &78.1&82.3&\underline{92.4}  \\
$^{\dag}$MRM~\cite{zhou2023mrm}           &\underline{83.6}&\underline{88.3}&92.1 \\
\rowcolor[gray]{0.9} \algname (Ours)      &\textbf{84.6}&\textbf{88.7}&\textbf{92.6} \\ \bottomrule
\end{tabular}
\vspace{-3mm}
\label{table:ffa-seg}
\end{table}

\subsection{Ablation Study and Analysis}

\bheading{Advantages of context distillation} The first row in Table~\ref{table:ab_component} is trained by basic MIM and MLM. When we use the distilled easy-comprehensible report, we observe a performance improvement of 3.5\% on COVIDx~\cite{wang2020covid} and 0.7\% on ChestX-ray14~\cite{wang2017chestx} while using 1\% training data. 
Moreover, when we train \algname without the distilled report, notable performance degradation is observed, particularly on COVIDx and RSNA~\cite{rsna}, validating the significant role of \textit{entity-aware context distillation}.

\bheading{Effectiveness of descriptors masking} After constantly masking the critical descriptive words (as shown in Fig.~\ref{fig:main}) preceding the entities, substantial performance enhancements are observed across both coarse-grained and fine-grained tasks, exceeding our initial expectations (as shown in Table~\ref{table:ab_component}). Accordingly, it is confirmed that \textit{descriptors masking} can better leverage the supervised signals in medical reports. 

\bheading{Sensitive analysis of the re-balancing factor} In the performance increment between the \textit{3rd} and the \textit{4th} rows as well as the \textit{7th} row and the \textit{last} row of Table~\ref{table:ab_component}, we highlight the effectiveness of our proposed \textit{re-balancing factor}. 
The AUC score increases by 1.1\% and 1.0\% respectively on ChestX-ray14~\cite{wang2017chestx}, and the ACC shows sequential improvements of 3.5\% and 3.2\% when utilizing 1\% and 10\% of training data.
It is evident that the \textit{re-balancing factor} primarily contributes to the performance of classification tasks, and is also effective on fine-grained tasks.
We further provide hyper-parameter analysis of $\lambda_{neg}$ corresponding to different imbalance ratios between positive and negative expressions. As shown in Table~\ref{table:lambda}, setting $\lambda_{neg}$ to $0.05$ according to the imbalance ratio (20:1) leads to the best performance. These results strongly demonstrate the effectiveness of re-balanced factor and its capability of encouraging pre-training model to learn precise disease semantics.

\bheading{Performance of context-guided SR} As listed in the \textit{4th} row and the \textit{last} row of Table~\ref{table:ab_component}, employing \textit{context-guided SR} provides performance gains from 83.0\% to 86.3\% and 92.3\% to 95.0\% on COVIDx. Notably, on SIIM, the Dice score improves by 0.7\% and 0.9\%, mAP sees a notable increase of 3.7\% and 3.5\% on RSNA while using 1\% and 10\% ratio, 
which validates its effectiveness in learning informative features by generating detailed pathology vision context.
This underscores the efficacy of \textit{context-guided super resolution} in aiding pre-trained models to capture prior knowledge of visual texture structure, particularly enhancing performance in fine-grained visual downstream tasks.

\begin{figure*}[t]
    \centering
    \includegraphics[width=1\linewidth]{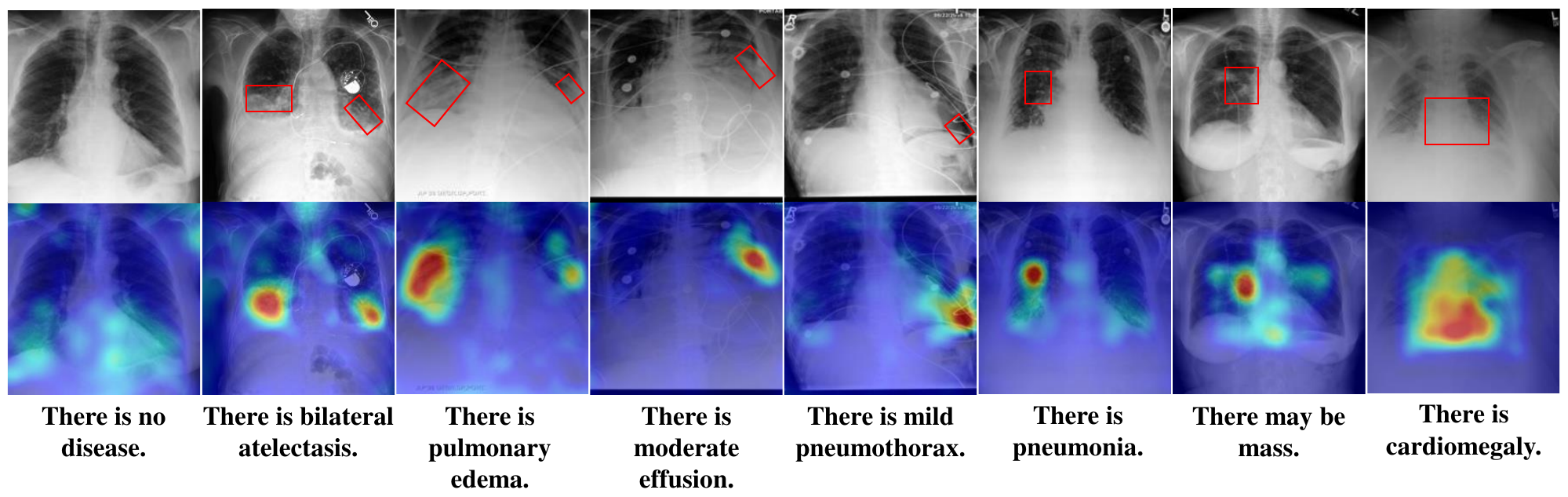}
    \vspace{-7mm}
    \caption{Cross-attention maps between diagnosis and vision patterns learned by local aggregated features in MSCF, which successfully localizes the pathology region of various diseases. }
    \label{fig:attention}
    \vspace{-5mm}
\end{figure*}

\bheading{Contributions of MSCF} We exclusively integrate the global vision feature $f_v^g$ and local aggregated feature $f_a^l$ to predict the masked words respectively. The results in Table~\ref{table:ab_component} validate that, only fusing $f_v^g$ decreases the performances on low-level downstream tasks (segmentation and detection), while the classification performance relies on well-trained $f_a^l$. It is confirmed that \textit{MSCF} successfully contributes to learning local and global representations and concurrently achieving SOTA performance on multi-scale downstream tasks.

\noindent{\textbf{Entity selection strategies} We conduct systematic ablation studies to evaluate the impact of \textit{entity selection strategies}. As illustrated in Table}~\ref{table:ab_text}, {we compare three entity selection strategies: (1) Top-20 highest-frequency entities, (2) Top-10\% frequency-percentile cutoff, and (3) our proposed method leveraging the Top-44 most frequent entities (as shown in Table}~\ref{table:entities} {). Compared to our method using 44 selected entities, strategies relying on the Top-20 entities or Top-10\% frequency thresholds exhibit diminished performance in modeling disease-related features, particularly under low-data fine-tuning scenarios.}

\noindent{\textbf{Text input composition} As our approach pre-trains vision encoder through the supervision of masked language modeling, text input composition may determine the performance of pre-trained models. For text input composition, we assess four paradigms: (1) original radiology reports alone, (2) ChatGPT-distilled reports alone, (3) hybrid inputs combining original reports with named entity recognition (NER)}~\cite{jain2021radgraph} {-based distilled reports, and (4) hybrid inputs combining original reports with ChatGPT-distilled reports. As shown in the last four rows of}~\ref{table:ab_text}{, using ChatGPT-distilled reports alone as input leads to significant performance degradation. While distilled reports capture disease existence and severity, they omit critical visual symptom descriptors, potentially impairing the model’s ability to localize fine-grained anatomical features. Hybrid inputs combining original and distilled reports yield more robust performance, as they balance structured context with raw textual detail.}

{We conduct further ablation study on context distillation method using the structured template ``[entity\_1] has [entity\_2]" with Radgraph-based NER annotations}~\cite{jain2021radgraph}{. Comparing the last two rows of Table}~\ref{table:ab_text}{, NER-based distillation method, though effective for entity presence/absence labeling, is less adept at capturing nuanced clinical relationships in complex reports. This limitation is evidenced by the limited improvement in multi-scale context-aware tasks, especially in fine-grained anatomical location-related tasks (e.g., segmentation and detection). In contrast, ChatGPT-based distillation method preserves nuanced clinical descriptors such as disease severity, anatomical location, and temporal progression (Table}~\ref{table:distill_example}){. Quantitative results demonstrate that our approach more effectively models disease severity and anatomical interactions, strongly supporting our hypothesis that ChatGPT-driven context distillation better preserves implicit clinical knowledge.}

\subsection{Representation analysis}
As the attention map of the local vision feature and the corresponding distilled reports visualized in Fig.~\ref{fig:attention}, the text modality of \algname captures the correct pathological regions of lung collapse (marked by red boxes) through cross-modal attention with the visual modality. Furthermore, there is no particularly disease-sensitive cross-attention for healthy images. These results support the ability of the \textit{MSCF} to learn accurate and discriminative fine-grained representations, which is crucial for improving the performance of local downstream tasks.

\section{Impact, Limitation, and Conclusion}

{In this paper, we propose a novel pre-training method which extracts representative and multi-scale features from complex and imbalanced medical reports. Our method is a cohesive integration of four modules, which contribute from different perspectives with close collaboration, emphasizing an entity-centered and context-aware design.} We confront the linguistic challenge of complex medical reports by distilling precise entity-centered knowledge using a large language model.
Subsequently, the catastrophic context-imbalanced issue is addressed by proposing a novel re-balancing factor, then entity-centered descriptors masking is proposed to strengthen the entity-centered context for masked language modeling.
Furthermore, a context-guided super-resolution learns to capture the critical pathology vision features with entity-awareness.
In addition, multi-scale context fusion helps to optimize informative global and local representations simultaneously.
These four readily accessible modules form the core of our simple-yet-effective general Med-VLP framework, referred to as \algname, whose significant superior performance is validated on various public datasets across CXRs and fundoscopies on downstream tasks, including classification, segmentation, and detection.  
\algname facilitates the development of a robust medical foundation model, ultimately alleviating the workload of experts. 
However, the absence of explicit alignment between text and vision modalities hinders the application in zero-shot setting.
In the future, we plan to incorporate contrastive learning to further empower \algname to function as an expert doctor without any manual annotation.

\section{Acknowledgments}
This work was supported by National Natural Science Foundation of China under Grant 62271465, Suzhou Basic Research Program under Grant SYG202338, and Open Fund Project of Guangdong Academy of Medical Sciences, China (No. YKY-KF202206).

\bibliographystyle{model2-names.bst}\biboptions{authoryear}
\bibliography{main}

\end{document}